\documentclass[runningheads]{llncs}

\usepackage[final,year=2026]{ricloc}

\usepackage{riclocabbrv}

\usepackage{graphicx}
\usepackage{booktabs}
\usepackage{amsmath}
\usepackage{amssymb}
\usepackage{bm}
\usepackage{multirow}
\usepackage{xcolor}
\usepackage{xspace}
\usepackage[accsupp]{axessibility}

\usepackage[breaklinks,colorlinks,citecolor=riclocblue]{hyperref}
\usepackage[capitalize]{cleveref}

\newcommand{\transp}{^{\top}}
\newcommand{\SOthree}{\ensuremath{\mathrm{SO}(3)}}
\newcommand{\SEthree}{\ensuremath{\mathrm{SE}(3)}}
\newcommand{\Simthree}{\ensuremath{\mathrm{Sim}(3)}}
\newcommand{\Real}{\ensuremath{\mathbb{R}}}

\newcommand{\sscale}{\ensuremath{s}}

\newcommand{\Rqi}{\ensuremath{R_{q,i}}}
\newcommand{\Cqi}{\ensuremath{C_{q,i}}}
\newcommand{\Rcons}{\ensuremath{R_{\mathrm{cons}}}}
\newcommand{\Ccons}{\ensuremath{C_{\mathrm{cons}}}}

\newcommand{\sigcons}{\ensuremath{\sigma_{\mathrm{cons}}}}
\newcommand{\sigdisp}{\ensuremath{\sigma_{\mathrm{disp}}}}
\newcommand{\sigjoint}{\ensuremath{\sigma_{\mathrm{joint}}}}
\newcommand{\Calign}{\ensuremath{C_{\mathrm{align}}}}

\newcommand{\method}{RIC-Loc\xspace}

\DeclareMathOperator*{\argmin}{arg\,min}

\newcommand{\ArxivEqualMark}{\textsuperscript{*}}
\newcommand{\ArxivCorrespondingMark}{\textsuperscript{\ensuremath{\dagger}}}

\newcommand{\ArxivAuthors}{%
  Wonseok Kang\inst{1}\ArxivEqualMark \and
  Jaehyun Kim\inst{2}\ArxivEqualMark \and
  Jeongmin Lee\inst{1} \and
  Tae-Wan Kim\inst{2}\ArxivCorrespondingMark%
}

\newcommand{\ArxivAuthorRunning}{W. Kang et al.}

\newcommand{\ArxivInstitutes}{%
  DeepFine Co., Ltd., Seoul, South Korea\\
  \email{eoid361@deepfine.com, jm.lee@deepfine.com}
  \and
  Seoul National University, Department of Naval Architecture and Ocean Engineering, Seoul, South Korea\\
  \email{jaedalong@snu.ac.kr, taewan@snu.ac.kr}\\[2pt]
  \ArxivEqualMark Equal contribution. \quad
  \ArxivCorrespondingMark Corresponding author.%
}

\newcommand{\MainCref}[1]{\cref{#1}}
\newcommand{\MainRef}[1]{\ref{#1}}

\begin{document}

\title{Reference-Induced Consensus for Selective\\Posed-Reference Visual Localization}
\titlerunning{Reference-Induced Consensus for Posed-Reference Localization}

\author{\ArxivAuthors}
\authorrunning{\ArxivAuthorRunning}
\institute{\ArxivInstitutes}

\maketitle

\begin{abstract}
We present \method{} (Reference-Induced Consensus localization), a scene-training-free posed-reference localizer that is SfM-point-map-free in its main estimator: it uses known reference poses, but not precomputed SfM 3D map points, query-to-map 2D--3D matches, or query-to-map PnP. A frozen VGGT pass predicts local camera poses, depth, and query-reference tracks for a query and selected references. Each reference induces one map-frame \SEthree{} query-pose hypothesis, robust consensus estimates the pose, and the preserved hypothesis structure yields two reliability scores: spatial dispersion and a track-conditioned covariance score. On the covariance-eligible set, the two scores are complementary for held-out, ground-truth-free failure detection across indoor, outdoor, and large-scale low-texture benchmarks: the joint policy is strongest in textured scenes and the covariance score in the low-texture regime, and the hypothesis-derived scores consistently outperform the standard retrieval-score gap and random rankings. Without per-scene training the consensus estimator remains accurate---competitive with structure-based localization indoors and improving over a comparable feed-forward baseline---giving an effective selective operating regime for posed-reference localization. Code is available at \url{https://github.com/SNU-DLLAB/ric_loc}.
\end{abstract}


\section{Introduction}\label{sec:intro}

\begin{figure}[t]
  \centering
  \includegraphics[width=\linewidth]{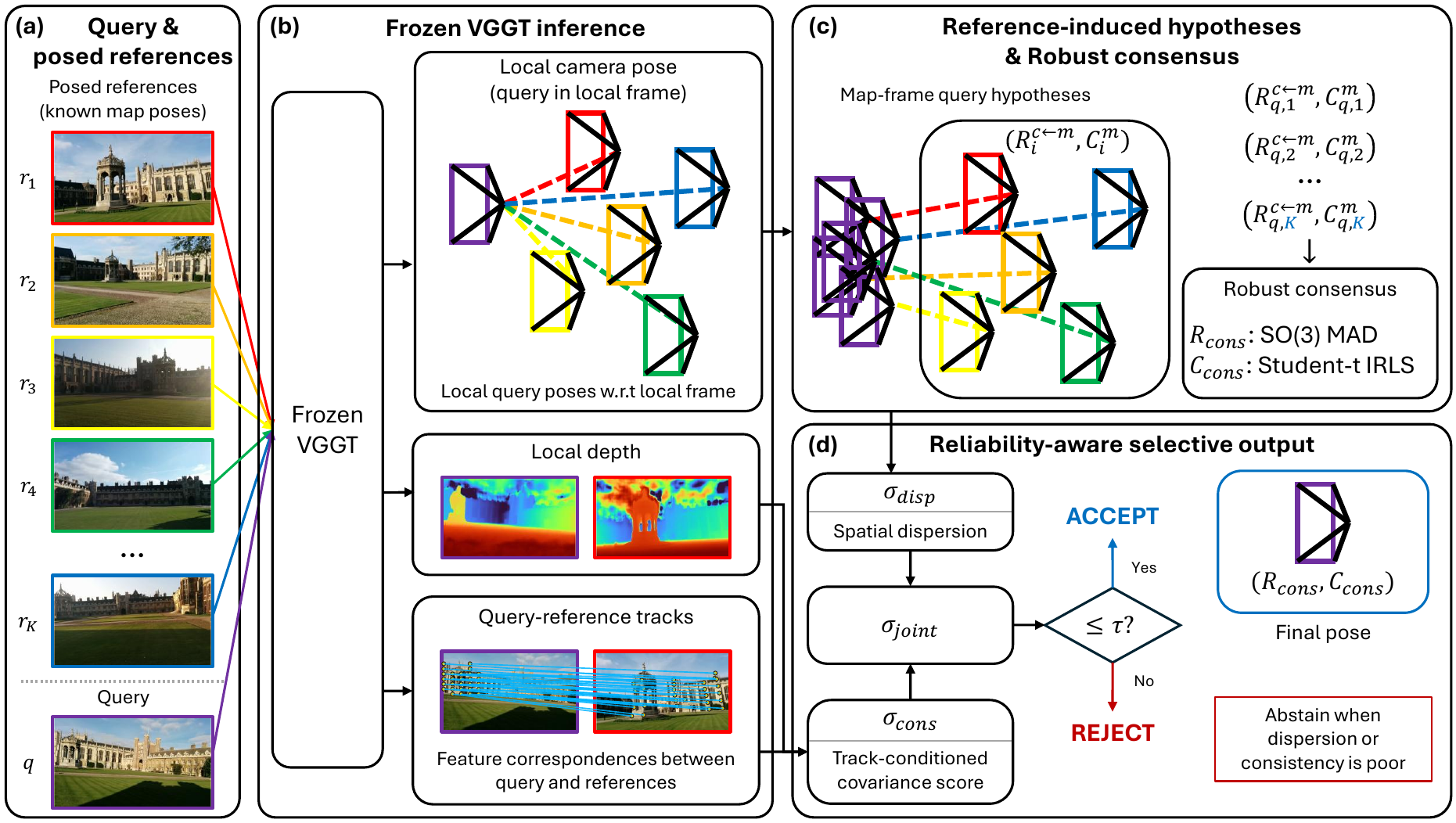}
  \caption{Selective posed-reference localization with \method{}. A single frozen VGGT pass
  over a query and its $K$ posed references predicts local poses, depth, and tracks; each reference
  lifts the query into the map frame as one \SEthree{} hypothesis, and robust
  $\SOthree{}\times\Real^3$ consensus produces $(\Rcons,\Ccons)$. The preserved structure yields
  dispersion \sigdisp{} and a track-conditioned covariance score \sigcons{}; their held-out-normalized
  fusion \sigjoint{} ranks failures and gates the pose.}
  \label{fig:pipeline}
\end{figure}

Feed-forward multi-view geometry models such as VGGT~\cite{vggt2025} and DUSt3R~\cite{dust3r2024}
predict local geometry, camera poses, and correspondences in a single forward pass. Their outputs are
not directly grounded in the coordinate frame of a deployed map.
Posed references provide that grounding: reference selection defines the hypothesis set, and each
reference can lift the query from the feed-forward local frame into a map-frame \SEthree{}
hypothesis.

Localization also needs rejection. Returning no pose delays a downstream task, but accepting a
wrong pose can seed tracking, map updates, or downstream AR and robot actions from a false global state. We
therefore treat pose estimation and rejection as coupled outputs of the same reference-induced
hypothesis structure. Collapsing selected references into a single aligned pose can discard
disagreement and track-support patterns that are informative for failure detection.

We present \method{} (Reference-Induced Consensus localization; \cref{fig:pipeline}), a scene-training-free posed-reference localizer. A frozen VGGT processes a
query and selected posed references, producing local poses, depth, and query-reference tracks. Each
reference induces one query-pose hypothesis $(\Rqi,\Cqi)$ in the map frame; robust
$\SOthree{}\times\Real^3$ consensus estimates $(\Rcons,\Ccons)$. The preserved hypotheses also
yield two reliability scores. The spatial-dispersion score \sigdisp{} is the root-mean-square spread
of the lifted center hypotheses about \Ccons, whereas the track-conditioned covariance score
\sigcons{} uses VGGT track support and local observability propagated through the lifting model rather
than hypothesis dispersion alone. Their normalized fusion is used for selective localization and
evaluated through risk--coverage curves.

We make three contributions.
\begin{enumerate}
  \item A hypothesis-preserving, SfM-point-map-free posed-reference scheme that induces one
    map-frame \SEthree{} query-pose hypothesis per reference without precomputed SfM map points,
    query-to-map 2D--3D matching, or PnP.
  \item Robust consensus localization over these hypotheses with no per-scene model training.
  \item Complementary reliability scores from the preserved hypothesis structure, evaluated by
    failure-detection AUROC and risk--coverage.
\end{enumerate}

We validate the consensus pose and the joint selective policy on indoor, outdoor, and large-scale
indoor benchmarks, and show that the lift-and-consensus procedure is not specific to VGGT, transferring to
other feed-forward front-ends under the identical posed-reference protocol.

\section{Related Work}
\label{sec:related}

Feed-forward multi-view geometry models predict geometry and cameras without per-scene optimization:
DUSt3R~\cite{dust3r2024} regresses dense pointmaps for an image pair, MASt3R~\cite{mast3r2024} adds a
metric matching head, and VGGT~\cite{vggt2025} predicts poses, depth, point maps, and correspondences
for many views in one forward pass, building on self-supervised transformer features~\cite{dinov2_2024}.
We use VGGT as a frozen local-geometry front-end for a query and its posed references; its
camera output is not the final pose, which is instead grounded in the map frame by the known reference
poses.

A large body of learned localizers amortize localization into learned scene or map representations.
Absolute pose regression maps an image to a global pose---PoseNet~\cite{posenet2015} and its
uncertainty-aware~\cite{bayesianposenet2016} and geometry-aware~\cite{geomfactors2018}
variants---while map-relative regression~\cite{marepo2024} conditions a scene-agnostic regressor on a
scene-specific map representation. Scene-coordinate methods learn pixel-to-3D mapping and solve pose
with robust PnP, from DSAC's differentiable RANSAC~\cite{dsac2017} to the fast ACE~\cite{ace2023} and
larger-scale GLACE~\cite{glace2024}. These are accurate, but the deployed map is encoded in learned
scene-specific weights or map representations.

Structure-based pipelines instead match a query against an explicit SfM reconstruction~\cite{colmap2016}:
classical systems use prioritized 2D--3D matching~\cite{activesearch2016}, while hierarchical
localization~\cite{hloc2019} pairs global retrieval~\cite{netvlad2016} with local features and learned
matchers~\cite{superpoint2018,superglue2020,lightglue2023} to form 2D--3D matches for PnP;
PixLoc~\cite{pixloc2021} instead refines coarse poses by feature-metric alignment to a 3D model. Such
explicit-map systems are strong on outdoor, urban-driving, and large indoor localization
benchmarks~\cite{aachen2018,robotcar2017,inloc2018} but rely on precomputed 3D maps and query-to-map
matching or alignment.
Closer to us, query-reference relative-pose methods avoid an explicit query-to-map point map:
Reloc3r~\cite{reloc3r2025} regresses query-reference relative poses and aggregates the induced absolute
poses, and MicKey~\cite{mickey2024} predicts metric relative pose from learned correspondences.
Aggregating relative or reference-induced motions connects to motion and rotation
averaging~\cite{govindu2001,rotavg2013} and \Simthree{} alignment~\cite{umeyama1991}; our point
estimator belongs to this lift-and-average family, and we compare against its matched variants in
\cref{tab:matched}.

Localization also needs to know when to abstain. Uncertainty modeling ranges from Bayesian pose
regression~\cite{bayesianposenet2016} to selective prediction~\cite{elyaniv2010} and distribution-free
conformal prediction~\cite{angelopoulos2023gentle} and learn-then-test risk
control~\cite{learnthentest2025}. We do not regress a parametric posterior; reliability comes from the
preserved hypothesis structure---spatial dispersion of the lifted centers and a track-conditioned
covariance score---and gates failures without query ground truth.

\method{} thus occupies a distinct point in this space: it needs no per-scene training or
scene-specific model, and its main estimator uses no SfM point map, query-to-map matching, or PnP.
Unlike query-reference relative-pose methods---whether regression-based or correspondence-based
\cite{reloc3r2025,mickey2024}---it uses VGGT as a frozen local-geometry front-end rather than a
black-box pose predictor, and the reference-induced
map-frame hypotheses are preserved rather than collapsed into one aligned
pose. This design lets the same reference-indexed state yield both the consensus pose and two
ground-truth-free reliability scores, positioning \method{} as a hypothesis-preserving,
selectively-gated posed-reference localizer built on feed-forward multi-view geometry.

\section{Method}\label{sec:method}

\method takes a query image $q$, a COLMAP-format~\cite{colmap2016} posed-reference map (used only as a container for
calibrated reference images and their known poses; any sparse 3D points are ignored), global
retrieval descriptors, and a frozen VGGT~\cite{vggt2025}, which predicts query-local depth, tracks,
local geometry, and the intrinsics used for reprojection. We use
$m$ for the map frame, $v$ for VGGT's shared local frame, and $c$ for a camera frame. Rotations are written by direction:
$R^{a\leftarrow b}$ maps vectors from frame $b$ to frame $a$. For a reference image $i$, the known
map pose is
$(R_i^{c\leftarrow m},C_i^m)$, with the COLMAP world-to-camera convention
\begin{equation}\label{eq:colmap_convention}
  x_i^c = R_i^{c\leftarrow m}(x^m-C_i^m),
  \qquad
  C_i^m = -(R_i^{c\leftarrow m})^\top t_i^{c\leftarrow m}.
\end{equation}
VGGT predicts local poses $(R_i^{c\leftarrow v},C_i^v)$ and
$(R_q^{c\leftarrow v},C_q^v)$ in the shared frame $v$.

Localization is organized around the reference-induced map-frame \SEthree{} hypothesis set
\begin{equation}\label{eq:hyp_distribution}
  \mathcal{H}_q(S_q)
  =
  \left\{
    (R_{q,i}^{c\leftarrow m},C_{q,i}^m,w_i)
  \right\}_{i\in S_q},
\end{equation}
where $S_q$ is the active reference set and the subscript $(q,i)$ marks the query pose induced by
reference $i$. Lifting yields one hypothesis per reference,
robust consensus estimates the candidate pose, and the same reference-indexed structure with its
track support yields the reliability scores for rejection.

\subsection{Overview}\label{sec:pipeline}

\Cref{fig:pipeline} summarizes the pipeline. We retrieve a ranked pool $P_q$ of $L=20$
MegaLoc~\cite{megaloc2025} images and pass only its top $K=8$ subset $S_q$ to VGGT. Each active reference $i\in S_q$ carries a fixed initial weight
$w_i^{\mathrm{init}}$ combining VGGT track confidence with a retrieval prior (worse rank or larger
distance $\to$ smaller weight; Supplementary Sec.~\ref{sec:active_weights}). This frozen
$L{=}20$, $K{=}8$ setting is used throughout; selection uses no 3D map points or query ground truth.

A single frozen VGGT forward pass processes $[q,r_1,\ldots,r_K]$ with $r_i\in S_q$.
Its camera, depth, and track heads provide the local poses, query geometry, and query-reference
correspondences used below; the final pose comes only after reference-induced lifting and consensus.

We estimate a similarity transform from VGGT local reference centers to known map reference centers,
\begin{equation}\label{eq:align}
  C_i^m \approx sR_S^{m\leftarrow v}C_i^v+t_S^m,
\end{equation}
using $w_i^{\mathrm{init}}$-weighted Umeyama~\cite{umeyama1991} inside RANSAC~\cite{ransac1981}, followed by robust
refinement. This transform also defines the direct \Simthree{} alignment baseline
\begin{equation}\label{eq:align_pose}
  C_{\mathrm{align}}^m=sR_S^{m\leftarrow v}C_q^v+t_S^m,
  \qquad
  R_{\mathrm{align}}^{c\leftarrow m}
  =
  R_q^{c\leftarrow v}(R_S^{m\leftarrow v})^\top,
\end{equation}
used for comparison only: the proposed estimator keeps the shared scale $s$ but estimates the query
center from reference-induced hypotheses rather than taking $C_{\mathrm{align}}^m$.

\subsection{Reference-induced SE(3) hypothesis set}\label{sec:lift}

Each reference defines a reference-specific orientation anchor from VGGT's local frame to
the map frame,
\begin{equation}\label{eq:anchor}
  A_i^{m\leftarrow v}
  =
  (R_i^{c\leftarrow m})^\top R_i^{c\leftarrow v}.
\end{equation}
Using this anchor, reference $i$ induces one map-frame query-pose hypothesis with rotation
\begin{equation}\label{eq:lift_rot}
  R_{q,i}^{c\leftarrow m}
  =
  R_q^{c\leftarrow v}(A_i^{m\leftarrow v})^\top
  =
  R_q^{c\leftarrow v}(R_i^{c\leftarrow v})^\top R_i^{c\leftarrow m},
\end{equation}
and center
\begin{equation}\label{eq:lift_center}
  C_{q,i}^m
  =
  C_i^m+sA_i^{m\leftarrow v}(C_q^v-C_i^v).
\end{equation}
\Cref{eq:lift_rot} transfers the VGGT query-reference relative orientation through the known
reference rotation. \Cref{eq:lift_center} expresses the local query-reference displacement in the
map frame, scales it by the shared \Simthree{} scale, and anchors it at the known reference center.
The global \Simthree{} rotation $R_S^{m\leftarrow v}$ is not re-applied in
\cref{eq:lift_center}: the reference-specific anchor already carries reference $i$'s map orientation.
Thus the per-reference lift does not collapse to the direct alignment baseline of
\cref{eq:align_pose}.

Query-seeded VGGT tracks visible in both views provide local geometric support for each lifted
hypothesis. For reference $i$, let $\mathcal{T}_i$ be the retained tracks after validity, depth, and
visibility filtering. Let $x_{ij}$ be the reference observation, $X_{qj}^{c_q}$ the query-camera 3D
point from VGGT depth, and $\hat{x}_{ij}$ its projection into reference $i$ under the VGGT
query-to-reference transform $T_{i\leftarrow q}$. The reprojection uses VGGT-predicted intrinsics
mapped to image pixel coordinates. With residual $e_{ij}=x_{ij}-\hat{x}_{ij}$, the confidence-weighted
Gauss--Newton information
matrix is
\begin{equation}\label{eq:info_matrix}
  H_i
  =
  \sum_{j\in\mathcal{T}_i}
  J_{ij}^{\top}W_{ij}J_{ij},
  \qquad
  J_{ij}
  =
  \frac{\partial e_{ij}}{\partial \xi_i},
\end{equation}
where $\xi_i\in\mathfrak{se}(3)$ perturbs
$T_{i\leftarrow q}$ by left multiplication,
$\exp(\xi_i^\wedge)T_{i\leftarrow q}$. The weights $W_{ij}$ combine VGGT track confidence with a
robust MAD pixel scale. We use the damped inverse as a local observability approximation,
\begin{equation}\label{eq:relative_cov}
  \Sigma_{\mathrm{rel},i}
  \simeq
  (H_i+\lambda I)^{-1},
\end{equation}
using fixed conditioning safeguards. This damped inverse captures local observability around the
VGGT-predicted relative pose and is used for relative weighting and failure ranking. Define
\[
  u_i^m=A_i^{m\leftarrow v}(C_q^v-C_i^v).
\]
The first-order center-lift Jacobian is
\begin{equation}\label{eq:lift_jacobian}
  G_i
  =
  s
  \left[
    (R_i^{c\leftarrow m})^\top
    \quad
    -[u_i^m]_\times (R_i^{c\leftarrow m})^\top
  \right],
\end{equation}
and the lifted center covariance is
\begin{equation}\label{eq:lift_cov}
  \Sigma_{C,i}
  =
  G_i\Sigma_{\mathrm{rel},i}G_i^\top.
\end{equation}
Scale variance is estimated robustly from reference-center ratios and added once to the consensus
covariance as a common-mode term. The construction is a first-order factorized approximation of the
lifted-center covariance. \Cref{eq:lift_jacobian} follows by first-order differentiation of the lifted
map-frame displacement under the left perturbation of $T_{i\leftarrow q}$; see Supplementary
Sec.~\ref{sec:center_lift_jacobian}. References with too few usable tracks receive an inflated
isotropic covariance, so covariance-weighted consensus can down-weight them rather than discard
them. If fewer than two covariance-bearing hypotheses remain after this fallback, center estimation
falls back to scalar center consensus and \sigcons{} is unavailable.

\subsection{Robust SE(3) consensus}\label{sec:consensus}

Repeated structures, viewpoint gaps, and front-end degeneracy can produce outlying hypotheses. We
therefore estimate rotation and center with separate robust consensus procedures over the
reference-induced hypothesis set.

For rotation, let $q_i$ be the sign-consistent unit quaternion corresponding to
$R_{q,i}^{c\leftarrow m}$. We first compute a weighted chordal mean
\begin{equation}\label{eq:quat_mean_initial}
  A
  =
  \sum_{i\in S_q}w_i^{\mathrm{init}}q_iq_i^\top,
  \qquad
  q^{(0)}
  =
  \operatorname{eigvec}_{\max}(A),
\end{equation}
where $w_i^{\mathrm{init}}$ combines the MegaLoc retrieval prior with VGGT track confidence;
the fixed formula is listed in Supplementary Sec.~\ref{sec:active_weights} and uses no map-point
support. Geodesic residuals
\[
  e_i^R=d_{\mathrm{SO}(3)}(R_{q,i}^{c\leftarrow m},R(q^{(0)}))
\]
define a robust MAD inlier set
\begin{equation}\label{eq:rot_mad_threshold}
  \tau_R
  =
  \max(\tau_{\min},m_R+\kappa_R\operatorname{MAD}_R),
  \qquad
	  \mathcal{I}_R
	  =
	  \{i\in S_q:e_i^R\le\tau_R\},
\end{equation}
with $m_R,\operatorname{MAD}_R$ the median/MAD of $\{e_i^R\}$ and fixed $\tau_{\min},\kappa_R$. Recomputing the chordal mean on $\mathcal{I}_R$ yields
\begin{equation}\label{eq:rcons}
	  A_{\mathcal{I}_R}
	  =
	  \sum_{i\in\mathcal{I}_R}w_i^{\mathrm{init}}q_iq_i^\top,
	  \qquad
	  q_{\mathrm{cons}}
	  =
	  \operatorname{eigvec}_{\max}(A_{\mathcal{I}_R}),
  \qquad
  R_{\mathrm{cons}}=R(q_{\mathrm{cons}}).
\end{equation}

For center estimation, let $\mathcal{U}_q\subseteq S_q$ be the set of references with usable,
possibly inflated, center covariance. The rotation inlier mask $\mathcal{I}_R$ and center set
$\mathcal{U}_q$ are kept separate because rotation residuals and track-derived center covariances
provide different failure-ranking cues. The center objective is
\begin{equation}\label{eq:ccons_opt}
  C_{\mathrm{cons}}^m
  =
  \argmin_C
  \sum_{i\in\mathcal{U}_q}
  \rho\left(
    (C-C_{q,i}^m)^\top\Sigma_{C,i}^{-1}(C-C_{q,i}^m)
  \right).
\end{equation}
For the three-dimensional Student-$t$ loss,
\[
  \rho(d^2)
  =
  \frac{\nu+3}{2}\log(1+d^2/\nu).
\]
IRLS with fixed convergence settings uses
\begin{equation}\label{eq:student_t_weights}
  \omega_i
  =
  \frac{\nu+3}{\nu+m_i^2},
  \qquad
  m_i^2
  =
  (C_{\mathrm{cons}}^m-C_{q,i}^m)^\top
  \Sigma_{C,i}^{-1}
  (C_{\mathrm{cons}}^m-C_{q,i}^m),
\end{equation}
and yields
\begin{equation}\label{eq:ccons}
  C_{\mathrm{cons}}^m
  =
  \left(
    \sum_{i\in\mathcal{U}_q}
    \omega_i\Sigma_{C,i}^{-1}
  \right)^{-1}
  \left(
    \sum_{i\in\mathcal{U}_q}
    \omega_i\Sigma_{C,i}^{-1}C_{q,i}^m
  \right).
\end{equation}
This stage performs robust aggregation of the reference-induced hypotheses; we use it as a point
estimator and a ranking signal rather than a calibrated \SEthree{} posterior.
If $|\mathcal{U}_q|<2$, the estimator can still produce a scalar center-consensus fallback pose, but
the query is not used for covariance-score ranking. Its \sigjoint{} input is unavailable, so the
runtime gate rejects it and full-set analyses rank it as least reliable.

The estimator outputs the candidate pose $R_{\mathrm{final}}=R_{\mathrm{cons}}$,
$C_{\mathrm{final}}^m=C_{\mathrm{cons}}^m$, with
$t_{\mathrm{final}}^{c\leftarrow m}=-R_{\mathrm{final}}C_{\mathrm{final}}^m$.
The selective stage either accepts this candidate or rejects it; it does not replace it with another
pose.

\subsection{Reliability and selective localization}\label{sec:selective}

We extract two complementary reliability scores from the preserved hypotheses and their track support.

The covariance-derived reliability score \sigcons{} is based on the aggregate covariance approximation
\begin{equation}\label{eq:agg_cov}
	  \Sigma_{\mathrm{agg}}
	  =
	  \left(
    \sum_{i\in\mathcal{U}_q}
    \omega_i\Sigma_{C,i}^{-1}
  \right)^{-1}.
\end{equation}
We inflate it by the weighted Mahalanobis residual-consistency factor
\begin{equation}\label{eq:phi}
  \phi=\max(1,Q/\mathrm{dof}),\quad
  Q=\sum_{i\in\mathcal{U}_q}\omega_i m_i^2,\quad
  \mathrm{dof}=\max(3|\mathcal{U}_q|-3,1),
\end{equation}
and add the common-mode scale term:
\begin{equation}\label{eq:sigcons}
	  \Sigma_{\mathrm{cons}}
	  =
	  \phi\Sigma_{\mathrm{agg}}
	  +
  \mathrm{var}_s
  \frac{\partial C_{\mathrm{cons}}^m}{\partial s}
  \Big(\frac{\partial C_{\mathrm{cons}}^m}{\partial s}\Big)^{\top},
  \qquad
  \sigma_{\mathrm{cons}}
  =
  \sqrt{\lambda_{\max}(\Sigma_{\mathrm{cons}})}.
\end{equation}
The resulting \sigcons{} is a track-conditioned, covariance-derived reliability score for the center
consensus, computed from local reprojection geometry and hypothesis aggregation. It quantifies the weakest-direction
uncertainty of the center consensus and serves as a center-reliability ranking signal (calibration
analysis in Supplementary \cref{sec:supp_cov}).

The spatial-dispersion signal is the root-mean-square spread of the center hypotheses about the
consensus,
\begin{equation}\label{eq:sigdisp}
  \sigma_{\mathrm{disp}}
  =
  \sqrt{
    \frac{1}{|\mathcal{U}_q|}
    \sum_{i\in\mathcal{U}_q}
    \lVert C_{q,i}^m-C_{\mathrm{cons}}^m\rVert^2
  }.
\end{equation}
This measures physical disagreement among lifted centers and provides failure-ranking cues
associated with baseline, viewpoint, and triangulation degeneracy. The two signals are
complementary: \sigdisp{} measures spread of the lifted hypotheses, while \sigcons{} is a
track-conditioned covariance-derived score propagated through the lifting model.

To combine them across scenes with different scales and textures, we use held-out normalization:
we compute medians $\tilde{\sigma}_{\mathrm{cons}}$ and $\tilde{\sigma}_{\mathrm{disp}}$, and define
the joint reliability score
\begin{equation}\label{eq:sigjoint}
  \sigma_{\mathrm{joint}}(q)
  =
  \max\left(
    \frac{\sigma_{\mathrm{cons}}(q)}{\tilde{\sigma}_{\mathrm{cons}}},
    \frac{\sigma_{\mathrm{disp}}(q)}{\tilde{\sigma}_{\mathrm{disp}}}
  \right).
\end{equation}
If \sigcons{} is unavailable, \sigjoint{} is also unavailable and the localizer rejects the query.
Otherwise, the localizer accepts the pose when $\sigma_{\mathrm{joint}}(q)\le\tau_{\mathrm{joint}}$
and rejects it otherwise. We report risk--coverage trade-offs by sweeping $\tau_{\mathrm{joint}}$,
rather than committing to a single tuned threshold. The held-out medians align score scales across
scenes; the calibration scope is stated in the Limitations.

\section{Experiments}\label{sec:exp}

\subsection{Setup and protocol}\label{subsec:setup}

We evaluate on four benchmarks: 7-Scenes~\cite{sevenscenes2013} ($n=17{,}000$ indoor RGB-D queries), Cambridge Landmarks~\cite{posenet2015} ($n=1{,}918$ city-scale queries over all five scenes including \textsc{GreatCourt}), 12-Scenes~\cite{twelvescenes2016} ($n=5{,}641$ queries over all twelve scenes), and NAVER Gangnam Station~\cite{naver2021} (B1: $2{,}620$ queries; B2: $916$ queries, with low-texture repetitive structure). We report median translation/rotation error and strict success at $5\,\text{cm}/5^{\circ}$ for 7-Scenes and 12-Scenes, $0.25\,\text{m}/2^{\circ}$ for Cambridge and NAVER, plus the three NAVER tiers. Unless otherwise stated, AUROC and risk--coverage use strict failure, meaning not strict-correct; catastrophic rate is a secondary safety metric, with catastrophic failure defined as $e_t>5\,\text{m}$ or $e_R>20^{\circ}$ (tightened to $0.5\,\text{m}/10^{\circ}$ on 7-Scenes).

\method{} and the feed-forward baseline Reloc3r-512~\cite{reloc3r2025} use the same frozen $L{=}20$, $K{=}8$ reference protocol with fixed hyperparameters (\cref{sec:pipeline}; full values in Supplementary \cref{sec:implementation_details}); hloc~\cite{hloc2019} (SuperPoint+LightGlue) is a structure-based context baseline with its own SfM point map and 2D--3D/PnP pipeline. Fusion normalization (\cref{eq:sigjoint}) uses held-out queries---leave-one-scene-out on 7-Scenes/Cambridge and cross-building on NAVER. We report selective metrics on 7-Scenes, Cambridge, and NAVER, with 12-Scenes as an additional accuracy benchmark; selective curves rank covariance-eligible queries ($47.3\%$ and $74.7\%$ of NAVER-B2/B1, $100\%$ and $98.8\%$ on 7-Scenes/Cambridge); full-set behavior (ineligible queries treated as least reliable or as failures) is reported separately.

\subsection{Selective localization}\label{subsec:selective_loc}

The objective is to rank failures before returning the candidate pose, using the dispersion and covariance scores from the preserved hypotheses under a unified joint policy.

\Cref{tab:rejecthead2head} measures how well each ground-truth-free signal separates strict failures
from successes: we compare $\sigcons$, $\sigdisp$, and their fusion $\sigjoint$ (\cref{eq:sigjoint})
against scalar center dispersion, Sim(3) conditioning, the top-1$-$top-2 retrieval-score gap, and a
random baseline. The retrieval gap is place-recognition confidence and does not affect \method{}'s
pose or scoring (Supplementary Sec.~\ref{sec:retrieval_gap}); held-out fusion medians give AUROC
within $0.5$\,pp of in-sample, so the ranking is robust to score-scale fitting.

\begin{table}[t]
\centering
\caption{Ground-truth-free strict-failure detection (AUROC; failure $=$ not strict-correct) on the
covariance-eligible set; $\sigjoint$ is the held-out-normalized max fusion of $\sigcons$ and
$\sigdisp$ and the last row is the eligible fraction. Best per column in bold.}
\label{tab:rejecthead2head}
\setlength{\tabcolsep}{4pt}
\resizebox{\columnwidth}{!}{%
\begin{tabular}{lcccc}
\toprule
Confidence signal & 7-Scenes & Cambridge & NAVER-B2 & NAVER-B1 \\
\midrule
\sigjoint{} (\method, median-norm.\ fusion) & \textbf{0.822} & \textbf{0.767} & 0.791 & 0.781 \\
\sigcons{} (\method, covariance score)  & 0.778 & 0.764 & \textbf{0.806} & \textbf{0.813} \\
\sigdisp{} (\method, spatial dispersion)    & 0.762 & 0.741 & 0.618 & 0.692 \\
\midrule
center dispersion (scalar)              & 0.756 & 0.751 & 0.650 & 0.717 \\
Sim(3) conditioning                     & 0.638 & 0.715 & 0.653 & 0.732 \\
retrieval-score gap (top-1$-$top-2)     & 0.529 & 0.540 & 0.516 & 0.546 \\
random baseline                         & 0.500 & 0.500 & 0.500 & 0.500 \\
\midrule
covariance-eligible (\%)                & 100 & 98.8 & 47.3 & 74.7 \\
\bottomrule
\end{tabular}}
\end{table}

The two hypothesis-derived signals are complementary: the joint score is strongest on 7-Scenes and
Cambridge ($0.822/0.767$), while \sigcons{} leads in the low-texture NAVER corridors ($0.806/0.813$).
Because both signals are preserved, the gate defaults to the joint score, with \sigcons{} the
single-signal choice in the low-texture regime; a paired bootstrap of the joint-vs-\sigcons{} difference is in
\cref{tab:significance}. All hypothesis-derived
scores outperform the retrieval gap and random on every split; the retrieval gap is near-random (AUROC
$0.52$--$0.55$), confirming retrieval confidence does not predict pose failure. A naive
consensus-inlier-count baseline is likewise near-random (Supplementary \cref{tab:supp_inlier}).

\begin{figure}[t]
  \centering
  \includegraphics[width=\linewidth]{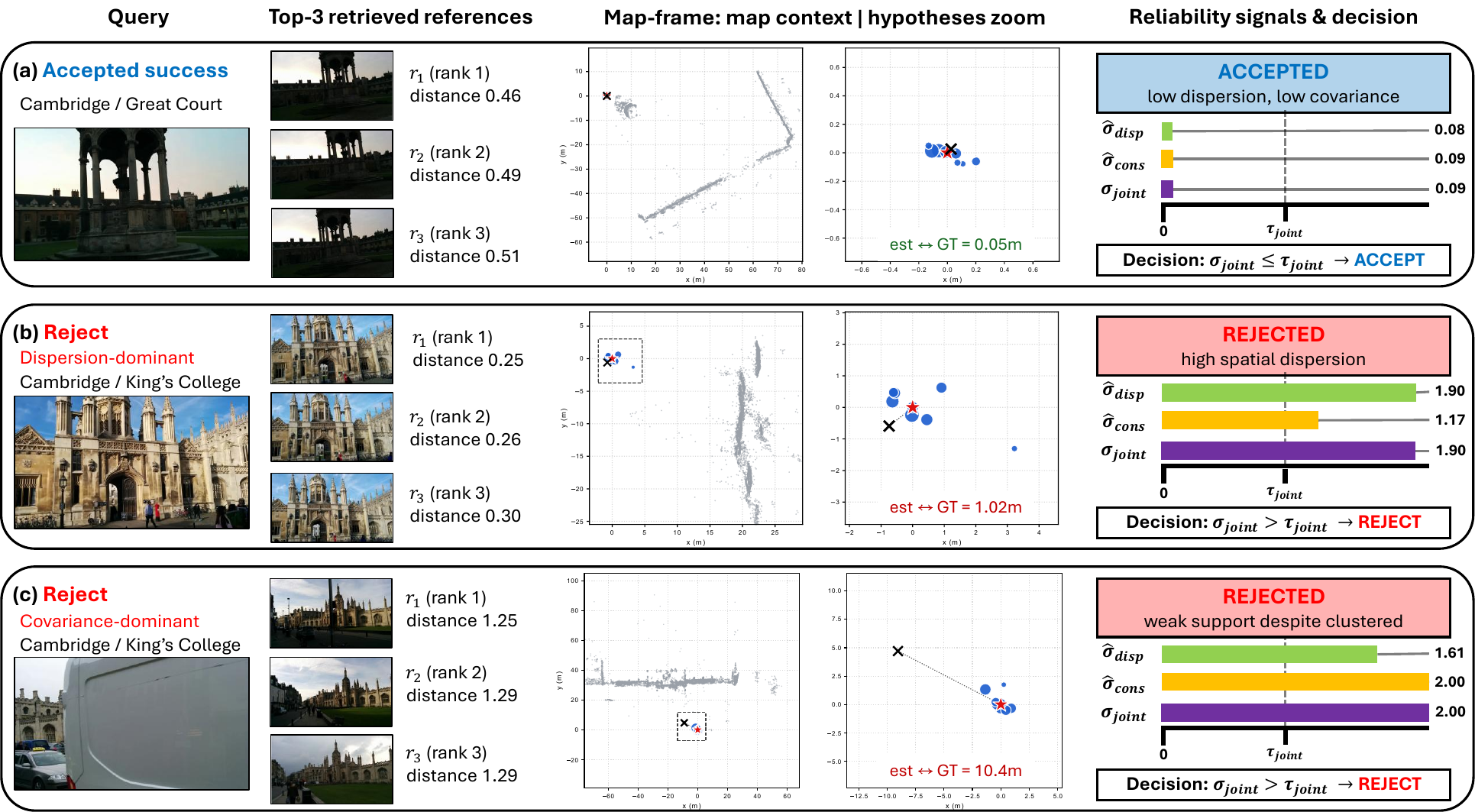}
  \caption{Reference-induced hypotheses and selective decisions. Each row is one query (top-3
  of $K{=}8$ references shown). Map-frame panels plot the lifted centers $C_{q,i}^m$ with
  per-reference covariance ellipses $\Sigma_{C,i}$; the right column reports the normalized scores
  $\hat{\sigma}_{\mathrm{disp}},\hat{\sigma}_{\mathrm{cons}}$ (each over its held-out median) and the gate $\sigma_{\mathrm{joint}}$.
  Rows: accepted success, spread-dominant rejection, covariance-dominant rejection. Ellipses are
  approximate; ground truth is for visualization only.}
  \label{fig:selective_accept_reject}
\end{figure}

\Cref{fig:selective_accept_reject} visualizes this complementarity through representative joint-gate decisions: one rejected case is dominated by spatial spread, while another is dominated by the track-conditioned covariance score despite a more clustered hypothesis pattern.

\begin{figure}[t]
  \centering
  \includegraphics[width=0.44\linewidth]{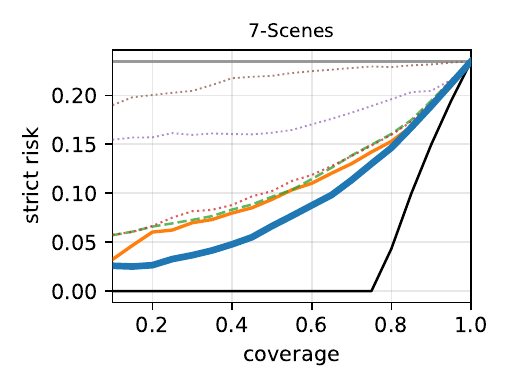}\hfill
  \includegraphics[width=0.44\linewidth]{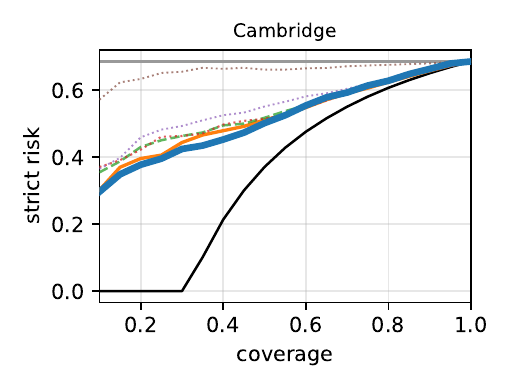}\\[1pt]
  \includegraphics[width=0.44\linewidth]{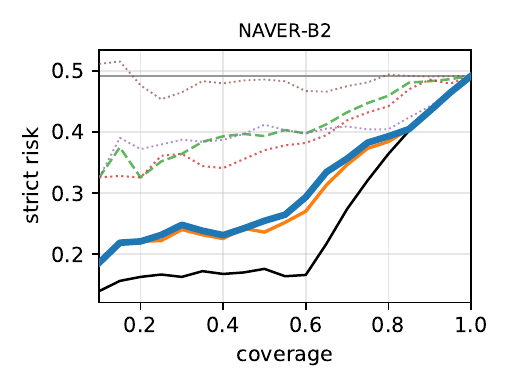}\hfill
  \includegraphics[width=0.44\linewidth]{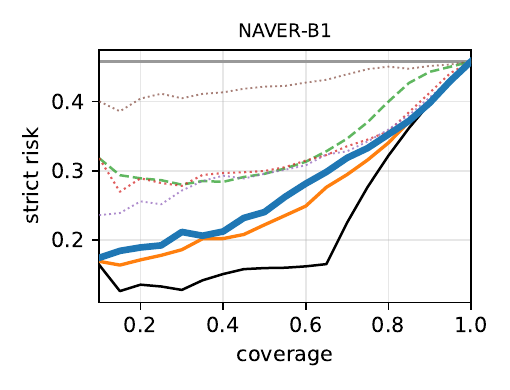}\\[1pt]
  \includegraphics[width=0.9\linewidth]{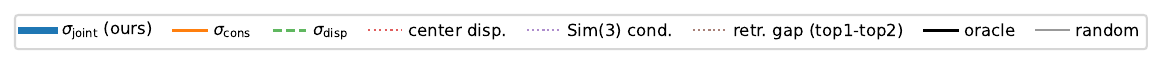}
  \caption{Risk--coverage by signal on the covariance-eligible set: accepted-set strict risk vs.\
  coverage (lower is better), queries accepted in increasing score order. \sigjoint{} (thick) and
  \sigcons{} run below the scalar and retrieval baselines over the operating range, with \sigjoint{}
  strongest on 7-Scenes/Cambridge and \sigcons{} in the low-texture NAVER buildings
  (cf.\ \cref{tab:rejecthead2head}). The oracle ranks by true translation error; random is the base
  failure rate.}
  \label{fig:riskcov}
\end{figure}

We then sweep the accepted set by lowest \sigjoint{} in \cref{tab:riskcoverage} and \cref{fig:riskcov}.
At the reported coverages, rejection lowers strict risk and catastrophic rate on every selective-evaluation split
(7-Scenes strict-risk $0.235\!\to\!0.146$ at $0.80$ coverage; NAVER catastrophic B2
$0.256\!\to\!0.133$, B1 $0.145\!\to\!0.017$). AURC improves over random
($0.087/0.489/0.298/0.270$ vs.\ $0.235/0.685/0.492/0.458$), with the same joint-vs-\sigcons{}
split across scenes as in \cref{tab:rejecthead2head}.

\begin{table}[t]
\centering
\caption{Risk--coverage under ground-truth-free joint selection on the covariance-eligible set
(accepted-set strict risk / catastrophic rate, lower better; queries accepted in increasing
$\sigjoint$ order). AURC is the strict-risk--coverage area; random/oracle are references, not
deployable baselines.}
\label{tab:riskcoverage}
\setlength{\tabcolsep}{6pt}
\resizebox{\columnwidth}{!}{%
\begin{tabular}{lcccc}
\toprule
Coverage & 7-Scenes (risk/cat.) & Cambridge (risk/cat.) & NAVER-B2 (risk/cat.) & NAVER-B1 (risk/cat.) \\
\midrule
Cov 1.0 (no rejection)  & 0.235 / 0.002 & 0.685 / 0.012 & 0.492 / 0.256 & 0.458 / 0.145 \\
Cov 0.8                 & 0.146 / 0.000 & 0.628 / 0.000 & 0.393 / 0.133 & 0.353 / 0.017 \\
Cov 0.7                 & 0.113 / 0.000 & 0.593 / 0.000 & 0.356 / 0.106 & 0.318 / 0.005 \\
Cov 0.5                 & 0.066 / 0.000 & 0.501 / 0.000 & 0.255 / 0.028 & 0.240 / 0.003 \\
\midrule
AURC $\sigjoint$ (\method) & 0.087 & 0.489 & 0.298 & 0.270 \\
AURC random / oracle & 0.235 / 0.030 & 0.685 / 0.322 & 0.492 / 0.148 & 0.458 / 0.126 \\
\bottomrule
\end{tabular}}
\end{table}

The preserved hypotheses also outperform black-box confidence: on NAVER the covariance score ranks
catastrophic failures more accurately than Reloc3r~\cite{reloc3r2025} self-agreement
(catastrophic-failure AUROC $0.92/0.97$ vs.\ $0.88/0.80$ on B2/B1).

\paragraph{Full-set selective behavior.} The selective gains hold over the entire query stream:
ranking the full query set with ineligible queries ($52.7\%/25.3\%$ of NAVER-B2/B1) placed
least-reliable raises full-set \sigcons{} AUROC to $0.941/0.893$ (B2/B1) and cuts full-set strict risk
at $0.5$ coverage ($0.760\!\to\!0.520$ B2, $0.595\!\to\!0.280$ B1). The NAVER gain thus goes beyond the
eligibility filter: \sigcons{} exceeds scalar dispersion among eligible queries and a gate-only ranker
over the full stream (NAVER-B1 $0.893$ vs.\ $0.713$ AUROC; paired bootstrap, Supplementary
\cref{tab:fullstream}). We report the within-eligible numbers of \cref{tab:rejecthead2head} as
the deployable operating point.

\subsection{Accuracy on benchmarks}\label{subsec:accuracy}

\Cref{tab:benchmark} reports accuracy on 7-Scenes, Cambridge, and 12-Scenes. On 7-Scenes, \method has the lowest
median translation ($2.9$\,cm) and rotation ($0.83^{\circ}$) among the compared
methods and the highest strict success ($0.765$, above structure-based hloc $0.662$ and Reloc3r-512
$0.670$), without per-scene training. On Cambridge, rotation is precise ($0.40^{\circ}$, ahead of Reloc3r-512), and
against the like-for-like SfM-point-map-free Reloc3r-512 \method{} improves both translation ($40.0$
vs.\ $55.6$\,cm) and strict success ($0.311$ vs.\ $0.233$, full-set). The structure-based hloc reference remains stronger on outdoor translation
(discussed in the Limitations). On 12-Scenes, \method{} improves over the comparable
feed-forward Reloc3r-512 baseline ($0.957$ vs.\ $0.750$ strict), approaching the structure-based hloc
upper bound ($0.985$); per-scene results are in Supplementary \cref{tab:supp_12scenes}. Finally, the black-box VGGT baseline ($\Calign$, \cref{tab:matched}) reaches only
$0.674/0.211$ strict on 7-Scenes/Cambridge versus \method{}'s $0.765/0.315$,
isolating the gain from the hypothesis-preserving lift.

\begin{table}[t]
\centering
\caption{Accuracy on benchmarks: pooled median translation (cm) / rotation (deg) and strict success
($5\,\text{cm}/5^{\circ}$ on 7-Scenes and 12-Scenes, $0.25\,\text{m}/2^{\circ}$ on Cambridge; all
scenes). \method{} strict counts covariance-ineligible queries as failures (full set), matching the
always-localizing baselines. hloc is structure-based (its own SfM point map), shown as context; Reloc3r uses the same
references as \method. Best per row in bold.}
\label{tab:benchmark}
\setlength{\tabcolsep}{6pt}
\resizebox{\columnwidth}{!}{%
\begin{tabular}{ll ccc}
\toprule
Benchmark & metric & \method & hloc~\cite{hloc2019} & Reloc3r-512~\cite{reloc3r2025} \\
\midrule
7-Scenes~\cite{sevenscenes2013} & median $t$/$R$ & \textbf{2.9}/\textbf{0.83} & 3.6/1.19 & 3.6/0.93 \\
                                & strict         & \textbf{0.765} & 0.662 & 0.670 \\
\midrule
Cambridge~\cite{posenet2015}    & median $t$/$R$ & 40.0/0.40 & \textbf{15.6}/\textbf{0.21} & 55.6/0.57 \\
(5 scenes)                      & strict         & 0.311 & \textbf{0.688} & 0.233 \\
\midrule
12-Scenes~\cite{twelvescenes2016} & median $t$/$R$ & 1.3/0.51 & \textbf{0.9}/\textbf{0.42} & 2.3/0.70 \\
(12 scenes)                     & strict         & 0.957 & \textbf{0.985} & 0.750 \\
\bottomrule
\end{tabular}}
\end{table}
\subsection{Large-scale indoor evaluation}\label{subsec:naver}

We evaluate on the NAVER Gangnam Station large-scale indoor benchmark~\cite{naver2021}, which contains low-texture repetitive structure and large viewpoint changes. \Cref{tab:naver} compares \method{} with MASt3R~\cite{mast3r2024}, DUSt3R~\cite{dust3r2024}, and Reloc3r~\cite{reloc3r2025} using the same reference subsets to isolate the pose-estimation stage.

\method{} is strongest at the tight $0.25\,\text{m}/2^{\circ}$ threshold on both buildings (B2
$24.0\%$ vs.\ MASt3R $15.6\%$ and DUSt3R $6.8\%$; B1 $40.5\%$ vs.\ MASt3R $8.2\%$ and DUSt3R
$11.7\%$), pairing this tight-threshold accuracy with the strongest selective rejection under these
repetitive low-texture corridors. At the looser $1\,\text{m}/5^{\circ}$ tier, \method{}'s eligibility
gate declines part of the low-texture tail, so the always-emitting baselines can cover more of that
tail.

\begin{table}[t]
\centering
\caption{NAVER Gangnam Station full-set recall (\%) over all queries (failed/declined/covariance-ineligible
\method{} queries count as failures). Baselines use the same reference subsets and emit a pose per
query; *MASt3R uses its coarse stage. ``elig.'' is \method{}'s covariance-eligible fraction; medT/medR
are per-method medians over localized queries (\method{}'s covariance-eligible subset vs.\ the
baselines' full query set), not a common population.}
\label{tab:naver}
\setlength{\tabcolsep}{4pt}
\resizebox{\columnwidth}{!}{%
\begin{tabular}{llcccccc}
\toprule
Building & Method & elig.\ \% & 0.1m/1$^{\circ}$ & 0.25m/2$^{\circ}$ & 1m/5$^{\circ}$ & medT (loc) & medR (loc) \\
\midrule
\multirow{4}{*}{B2 (916q)}
 & \method{} (SfM-point-map-free) & 47.3 & 2.0 & 24.0 & 33.6 & 0.193m & 1.52$^{\circ}$ \\
 & MASt3R* (SfM-point-map-free refs) & 100  & 1.7 & 15.6 & 54.1 & 0.630m & 2.29$^{\circ}$ \\
 & DUSt3R (SfM-point-map-free refs) & 100  & 0.4 & 6.8  & 28.3 & 36m & 4.3$^{\circ}$ \\
 & Reloc3r (SfM-point-map-free refs) & 100  & 0.0 & 0.0  & 0.4  & 38m & 91$^{\circ}$ \\
\midrule
\multirow{4}{*}{B1 (2620q)}
 & \method{} (SfM-point-map-free) & 74.7 & 3.9 & 40.5 & 61.9 & 0.197m & 1.40$^{\circ}$ \\
 & MASt3R* (SfM-point-map-free refs) & 100  & 0.9 & 8.2  & 55.6 & 0.825m & 2.01$^{\circ}$ \\
 & DUSt3R (SfM-point-map-free refs) & 100  & 0.5 & 11.7 & 58.2 & 0.619m & 2.35$^{\circ}$ \\
 & Reloc3r (SfM-point-map-free refs) & 100  & 0.0 & 0.0  & 0.9  & 4.7m  & 86$^{\circ}$ \\
\bottomrule
\end{tabular}}
\end{table}

\subsection{Ablations}\label{subsec:ablation}

\paragraph{Estimator.} \Cref{tab:matched} fixes retrieval, references, and the single frozen VGGT pass,
holds rotation at $\Rcons$, and varies only the center estimator. The main gain comes from
replacing the direct Sim(3)-aligned query center $\Calign$ with per-reference lifted centers: strict
success improves from $0.674$ to $0.765$ on 7-Scenes, from $0.211$ to $0.315$ on Cambridge, and from
$0.353$ to $0.508$ on NAVER-B2 (covariance-eligible set; the NAVER-B2 gain persists full-set,
$16.7\%\!\to\!24.0\%$). The lift, not the aggregation, drives the indoor and outdoor gain. Covariance weighting is the operating
choice because it delivers the largest accuracy gain in the hardest, low-texture NAVER regime
($0.436\!\to\!0.508$) while trading under $2$\,pp against the other aggregators on 7-Scenes and
Cambridge ($0.765$--$0.784$/$0.315$--$0.326$), and it uniquely
yields the covariance-derived reliability score used in selective localization. With the selective
results (\cref{subsec:selective_loc}), the ablation supports the central claim: the lift improves the
estimator, and preserving the hypothesis structure enables rejection.

\begin{table}[t]
\centering
\caption{Matched estimator ablation. Same retrieval, references, and single frozen VGGT pass; all rows
share rotation $\Rcons$ and vary only the center estimator, except the black-box $\Calign$ row
(direct-alignment pose $R_{\mathrm{align}},\Calign$). Strict success at $5\,\text{cm}/5^{\circ}$
(7-Scenes) and $0.25\,\text{m}/2^{\circ}$ (Cambridge, NAVER-B2), over \method{}'s covariance-eligible set
(identical across rows, isolating the estimator); NAVER-B2 is thus eligible-subset accuracy
($0.508{\times}0.473$ gives the full-set $24.0\%$ of Tab.~\ref{tab:naver}) and Cambridge the
eligible-conditional $0.315$ (full-set $0.311$, Tab.~\ref{tab:benchmark}). $\dagger$: matched late-fusion
baselines. Bold $=$ best per column; bold row label $=$ the deployed estimator.}
\label{tab:matched}
\setlength{\tabcolsep}{6pt}
\resizebox{\columnwidth}{!}{%
\begin{tabular}{lccc l}
\toprule
Center estimator ($\Rcons$ fixed) & 7-Scenes & Cambridge & NAVER-B2 & isolates \\
\midrule
Direct Sim(3), black-box VGGT ($\Calign$)  & 0.674 & 0.211 & 0.353 & direct-alignment center \\
Plain lift-and-average$^\dagger$       & 0.769 & 0.320 & 0.411 & late-fusion baseline \\
Retrieval-weighted avg.$^\dagger$      & \textbf{0.784} & 0.322 & 0.436 & retrieval prior \\
Robust consensus only$^\dagger$        & 0.771 & \textbf{0.326} & 0.427 & robust averaging \\
\textbf{Robust\,$+$\,covariance ($\Ccons$)}     & 0.765 & 0.315 & \textbf{0.508} & covariance weighting \\
\bottomrule
\end{tabular}}
\end{table}

\paragraph{Backbone transfer.} As a transfer diagnostic rather than a primary benchmark comparison,
the same reference-induced lift-and-consensus procedure also improves DUSt3R and MASt3R raw
\Simthree{} outputs under the matched posed-reference protocol on every reported split
(Supplementary Tab.~\ref{tab:backbone}).

\paragraph{Efficiency.} At $K=8$, \method runs in $1.9$--$3.4$\,s per query across the evaluated
datasets (per-stage medians in Supplementary \cref{tab:supp_runtime}) on a single $24$\,GB
RTX 4090, with $13.1$\,GB peak memory and no per-scene training; accuracy and failure ranking are
stable across the reference count $K$ (Supplementary \cref{tab:supp_ksweep}).

\section{Conclusion}
\label{sec:conc}

We presented \method, a scene-training-free posed-reference localizer with an
SfM-point-map-free main estimator. Our formulation centers on reference-induced
hypothesis preservation. Reference selection determines the hypothesis set, with
each posed reference inducing one map-frame \SEthree{} pose hypothesis for the
query. Robust consensus then estimates the final pose, while the preserved
reference-indexed structure provides ground-truth-free reliability signals for
selective localization.

Experiments show that this lift-and-consensus design improves over matched
feed-forward baselines, is competitive on indoor posed-reference benchmarks, and
produces informative empirical risk--coverage trade-offs from hypothesis-derived
dispersion and covariance scores. The ablations indicate that the main gain comes
from replacing direct \Simthree{} alignment with reference-induced hypotheses, while
the preserved hypothesis structure supplies reliability cues that retrieval
confidence and scalar dispersion alone miss. These results support \method{} as a
hypothesis-preserving interface between feed-forward multi-view geometry, robust
map-frame pose consensus, and selective rejection.

\paragraph{Limitations and future work.} (i)~Outdoor accuracy. Without an explicit 3D map or
query-to-map PnP, outdoor translation trails structure-based SfM/PnP localization (Cambridge,
\cref{tab:benchmark}; still ahead of the like-for-like Reloc3r-512). Geometry-aware reference selection
within the posed-image database---for viewpoint diversity, predicted track support, and hypothesis
conditioning rather than fixed top-$K$ retrieval---is a natural next step. (ii)~Approximate uncertainty.
The track-conditioned covariance is first-order: its $K$ hypotheses share one VGGT pass yet are
aggregated as if independent, so it is over-confident in absolute scale ($\approx4$--$6\times$,
\cref{tab:cov_calib}) and omits rotation; we use \sigcons{} only as a monotone failure-ranking
signal, and a calibrated, rotation-aware version would add guaranteed risk control. (iii)~Coverage. On
low-texture repetitive scenes covariance eligibility can be limited (NAVER-B2 $47.3\%$), capping
deployable coverage; covariance-free fallback ranking for ineligible queries is an important direction.
(iv)~Cost and assumptions. Inference is dominated by the frozen VGGT pass ($1.9$--$3.4$\,s,
\cref{tab:supp_runtime}), heavier than retrieval-PnP; \method{} also needs held-out median scales
and assumes calibrated posed references.

\clearpage
\setcounter{section}{0}
\setcounter{table}{0}
\setcounter{figure}{0}
\setcounter{equation}{0}
\renewcommand{\thesection}{\Alph{section}}
\renewcommand{\thetable}{S\arabic{table}}
\renewcommand{\thefigure}{S\arabic{figure}}
\renewcommand{\theequation}{S\arabic{equation}}
\renewcommand{\theHsection}{supp.\arabic{section}}
\renewcommand{\theHtable}{supp.\arabic{table}}
\renewcommand{\theHfigure}{supp.\arabic{figure}}
\renewcommand{\theHequation}{supp.\arabic{equation}}

\begin{center}
{\Large\bfseries Supplementary Material\par}
\end{center}

\noindent This supplement records implementation details, active-reference weights, the center-lift
Jacobian, the retrieval-score-gap diagnostic, and backbone transfer. It then presents additional
covariance, selection, eligibility, significance, runtime, ranking-behavior, and per-scene analyses.
All reported numbers are computed from the same evaluation outputs used in the main paper.

\section{Implementation Details and Fixed Hyperparameters}
\label{sec:implementation_details}

Except for the explicitly labeled sensitivity sweep, the main-experiment hyperparameters below are
fixed across scenes and datasets (no per-scene tuning).
\begin{itemize}
  \item \textbf{Retrieval and active-set sizes:} MegaLoc~\cite{megaloc2025} retrieval pool $L=20$; active-reference count
  $K=8$ (the top-$K$ subset passed to VGGT). Sensitivity to $K$ is reported in \cref{sec:k_sensitivity}.
  \item \textbf{Initial active-reference weights:} the fixed $w_i^{\mathrm{init}}$ (with $\sigma_0=0.05$
  and the confidence/rank/distance clips) are given in full in \cref{sec:active_weights}.
  \item \textbf{Alignment:} RANSAC \Simthree{} (Umeyama, scale estimated): $2000$ iterations, inlier
  threshold $0.30$\,m, minimum $5$ inliers; on failure, a fallback pass uses $6000$ iterations,
  $0.40$\,m, and minimum $4$ inliers. The refit is a confidence-weighted Umeyama with IRLS robust
  reweighting. The alignment RANSAC seed is fixed at $42$.
  \item \textbf{Rotation consensus:} quaternion-chordal weighted mean (leading eigenvector of
  $\sum_i w_i\,q_i q_i^{\top}$); robust inlier gate at
  $\max(\tau_{\min},\mathrm{med}+\kappa_R\,1.4826\,\mathrm{MAD})$, with
  $\tau_{\min}=5^{\circ}$ and $\kappa_R=2.5$.
  \item \textbf{Robust center consensus:} Student-$t$ IRLS with $\nu=5$, at most $20$ iterations, and
  convergence tolerance $10^{-7}$.
  \item \textbf{Covariance construction:} information damping $\lambda=10^{-6}$; eigenvalues are clamped
  to $[10^{-8},10^{6}]$ when inverting the relative-pose information and to $[10^{-12},10^{12}]$ for the
  aggregate inverse-information approximation; the robust pixel-scale floor is
  $\sigma_{\mathrm{pix},\min}=1$\,px. A reference with fewer than $6$ tracks uses the inflated isotropic
  fallback $\Sigma_{C,i}=(10^{4})^2 I_3$. A query is covariance-eligible when the required track and
  reconstruction outputs are present, at least two covariance-bearing hypotheses remain after this
  fallback, and the resulting reliability score is finite. This covariance path applies no separate
  same-space eligibility gate; ineligible queries count as failures in full-set recall.
  \item \textbf{Scale uncertainty:} with $N_h$ retained covariance-bearing hypotheses,
  $\mathrm{var}_s=(1.4826\,\mathrm{MAD}(\{s_{ij}\}))^2/N_h$ for at least two valid ratios, and $0$
  otherwise, where $s_{ij}=\lVert C_i^m-C_j^m\rVert/\lVert C_i^v-C_j^v\rVert$ over reference pairs
  (no SfM 3D map points).
  \item \textbf{Selective normalization:} \sigjoint{} is the maximum of
  $\sigcons/\mathrm{med}(\sigcons)$ and $\sigdisp/\mathrm{med}(\sigdisp)$. The medians are estimated on
  held-out data: leave-one-scene-out on 7-Scenes and Cambridge, and cross-building on NAVER. A
  held-out-free rank-fusion variant closely matches the held-out joint on 7-Scenes/Cambridge, while
  \sigcons{} alone is stronger on the low-texture NAVER buildings; see \cref{sec:supp_openworld}.
  \item \textbf{Runtime:} per-query wall-clock on a single $24$\,GB GPU at batch size $1$ with a frozen
  VGGT (no gradients); median per-stage timings are reported in \cref{sec:supp_ci}.
\end{itemize}

\section{Sensitivity to the number of active references $K$}
\label{sec:k_sensitivity}

\Cref{tab:supp_ksweep} reports a sensitivity run over the active-reference count $K$ on 7-Scenes
($850$-query subset). The sweep also increases selection-pool capacities with $K$, so
it is not an isolated causal ablation of $K$. Point accuracy improves across the sweep
(median translation $3.08\!\to\!2.75$\,cm, strict success $0.748\!\to\!0.796$ over
$K{=}4\!\to\!16$), while rank-fused failure-detection AUROC remains in $[0.795,0.835]$. This indicates
that the observed reliability ranking is not tied to one reference-count setting. We use $K=8$ as
the fixed default for all main experiments.

\begin{table}[t]\centering
\caption{Reference-count sensitivity on 7-Scenes ($850$ queries); the recorded sweep increases the
selection-pool capacities with $K$. Median translation/rotation, strict success at
$5\,\text{cm}/5^{\circ}$, and ground-truth-free
failure-detection AUROC for $\sigcons$ and the rank-fused $\sigjoint$. Best per column in bold;
$K=8$ is the default used throughout the paper. These numbers use the $850$-query 7-Scenes subset and the held-out-free
rank-fused $\sigjoint$, so the $K{=}8$ row differs slightly from the full-set ($n=17{,}000$),
held-out-normalized headline of main-paper \MainCref{tab:benchmark}, \MainCref{tab:rejecthead2head}
($<0.4$\,pp strict, within the bootstrap CIs of \cref{tab:supp_ci}); the ordering and trends are
unchanged.}
\label{tab:supp_ksweep}
\setlength{\tabcolsep}{6pt}
\begin{tabular}{cccccc}
\toprule
$K$ & med $t$ (cm) & med $R$ ($^{\circ}$) & strict@$5\,\text{cm}/5^{\circ}$ & AUROC $\sigcons$ & AUROC $\sigjoint$ \\
\midrule
4  & 3.08 & 0.84 & 0.748 & \textbf{0.805} & 0.817 \\
6  & 3.07 & 0.85 & 0.759 & 0.761 & 0.795 \\
\textbf{8}  & 2.92 & 0.85 & 0.762 & 0.790 & 0.821 \\
12 & 2.87 & 0.80 & 0.793 & 0.745 & 0.805 \\
16 & \textbf{2.75} & \textbf{0.78} & \textbf{0.796} & 0.783 & \textbf{0.835} \\
\bottomrule
\end{tabular}
\end{table}

\section{Active-Reference Weights}
\label{sec:active_weights}

We retrieve a ranked candidate pool $P_q$ of $L=20$ map images with
MegaLoc. MegaLoc returns squared-L2 distances $d_i$ in ascending order, so smaller
values indicate better retrieval matches. Only the top $K=8$ subset $S_q\subset P_q$ is passed to
VGGT. For active reference $i\in S_q$, let $N_q=|P_q|$, $r_i$ be its retrieval rank,
$\bar c_i=\operatorname{clip}_{[0.05,1]}(c_i)$ its mean visible-track confidence, and
$d_{\min},d_{\max}$ the distance extrema over $P_q$. Define
$D_q=\max(d_{\max}-d_{\min},10^{-12})$ and $n_q=\max(N_q-1,1)$. The fixed initial weight is ($\sigma_0=0.05$)
\begin{align}
  \delta_i&=\frac{d_i-d_{\min}}{D_q},\qquad
  u_i=\operatorname{clip}_{[0,2]}\!\left(\sqrt{\frac{r_i-1}{n_q}}+
  \operatorname{clip}_{[0,1]}(\delta_i)\right),\\
  w_i^{\mathrm{init}}&=\left[\sigma_0^2+\sigma_0^2\frac{1-\bar c_i}{\bar c_i}
  +\sigma_0^2u_i^2\right]^{-1}.
\end{align}
Thus smaller MegaLoc distances and better ranks yield larger weights. Pool-level statistics only
parameterize weights for active references $i\in S_q$; unselected images do not pass through VGGT
and do not supply lifted hypotheses, track covariances, consensus summands, or reliability terms.
The weights use no SfM 3D map-point support.

\section{Center-Lift Jacobian Derivation}
\label{sec:center_lift_jacobian}

We derive the Jacobian used to propagate the local relative-pose approximation into the lifted
map-frame query center in \MainCref{eq:lift_jacobian}. The differentiated quantity is the lifted
displacement from reference center $C_i^m$ to query center $C_{q,i}^m$,
\[
  C_{q,i}^m-C_i^m
  =sA_i^{m\leftarrow v}(C_q^v-C_i^v)
  =s u_i^m,
  \qquad
  u_i^m=A_i^{m\leftarrow v}(C_q^v-C_i^v).
\]
The relative transform $T_{i\leftarrow q}$ is perturbed on the left as
$\exp(\xi_i^\wedge)T_{i\leftarrow q}$, with
$\xi_i=(\delta\rho_i^\top,\delta\theta_i^\top)^\top\in\mathfrak{se}(3)$. Thus
$\delta\rho_i$ and $\delta\theta_i$ are expressed in reference camera coordinates. Mapping them to
the map frame uses $(R_i^{c\leftarrow m})^\top$. To first order, the induced center change is
\[
  \delta C_{q,i}^m
  \simeq
  s(R_i^{c\leftarrow m})^\top\delta\rho_i
  -s[u_i^m]_\times(R_i^{c\leftarrow m})^\top\delta\theta_i
  =G_i\xi_i,
\]
where
\[
  G_i
  =s\left[
    (R_i^{c\leftarrow m})^\top
    \quad
    -[u_i^m]_\times(R_i^{c\leftarrow m})^\top
  \right]
  \in\mathbb{R}^{3\times6}.
\]
The first block maps a translational perturbation into the map frame. The second block follows from
the infinitesimal rotation of the lever arm: for a map-frame perturbation
$\delta\theta_i^m=(R_i^{c\leftarrow m})^\top\delta\theta_i$, the change is
$\delta\theta_i^m\times u_i^m=-[u_i^m]_\times\delta\theta_i^m$. This derivation drops terms of order
$\mathcal{O}(\|\xi_i\|^2)$ and is therefore a first-order approximation.

\section{Retrieval-Score Gap Diagnostic}
\label{sec:retrieval_gap}

The place-recognition confidence reported in main-paper \MainCref{tab:rejecthead2head} is the
standard top-1$-$top-2 MegaLoc gap: MegaLoc returns squared-L2 distances in ascending order, so the
gap is $g_q=d_2-d_1$ over the two best matches (larger $\Rightarrow$ more confident), scored with a
failure-oriented sign. Its failure-detection AUROC is near-random ($0.529/0.540/0.516/0.546$ on
7-Scenes/Cambridge/NAVER-B2/B1). An alternative pool-tail margin $d_{L-1}-d_L$ (separation at the
bottom of the pool) is likewise near-random ($0.545/0.561/0.466/0.524$), so retrieval-score separation
does not predict pose failure in either form. The diagnostic is computed separately from the ranked
$L=20$ pool and does not affect the \method{} pose estimate, $w_i^{\mathrm{init}}$, $\sigdisp$,
$\sigcons$, or $\sigjoint$.

\section{Backbone Transfer}
\label{sec:backbone_transfer}

Table~\ref{tab:backbone} reports a backbone-transfer diagnostic. We substitute
DUSt3R~\cite{dust3r2024} and MASt3R~\cite{mast3r2024} for VGGT while retaining the same references and
map-lift protocol. Here, raw denotes the plain \Simthree{} black-box output, whereas $+$\method{}
applies the reference-induced lift and robust consensus. Both variants use each backbone's
global-aligner poses. The matched results improve strict success for both backbones on all four
reported benchmark splits. They are not directly comparable to the native MASt3R pipeline in
main-paper Tab.~\MainRef{tab:naver}.

\begin{table}[t]
\centering
\caption{Backbone transfer. The lift-and-consensus procedure (\method) is applied with DUSt3R and
MASt3R substituted for VGGT under the same references and map-lift protocol; strict success, raw $=$
plain \Simthree{} black-box, $+$\method $=$ reference-induced lift and robust consensus. The recipe
improves these two backbones on every reported benchmark split in this matched protocol. Poses use each backbone's
global-aligner output, not the native MASt3R pipeline in main-paper \MainCref{tab:naver}; the two
tables are not directly comparable. NAVER strict success here is over the matched covariance-eligible
subset used for this diagnostic, not the full query set of main-paper \MainCref{tab:naver}, so the raw
entries differ from that table's full-set values (e.g.\ DUSt3R NAVER-B2 $0.032$ vs.\ $6.8\%$).}
\label{tab:backbone}
\setlength{\tabcolsep}{6pt}
\begin{tabular}{ll cccc}
\toprule
Backbone & & 7-Scenes & Cambridge & NAVER-B2 & NAVER-B1 \\
\midrule
\multirow{2}{*}{DUSt3R~\cite{dust3r2024}} & raw & 0.150 & 0.020 & 0.032 & 0.067 \\
 & $+$\method & \textbf{0.370} & \textbf{0.097} & \textbf{0.069} & \textbf{0.113} \\
\midrule
\multirow{2}{*}{MASt3R~\cite{mast3r2024}} & raw & 0.258 & 0.029 & 0.044 & 0.134 \\
 & $+$\method & \textbf{0.487} & \textbf{0.111} & \textbf{0.096} & \textbf{0.189} \\
\bottomrule
\end{tabular}
\end{table}

\section{Additional Track-Conditioned Covariance Details}
\label{sec:supp_cov}

Using the center-lift Jacobian derived in Supplementary Sec.~\ref{sec:center_lift_jacobian}, the
per-reference covariance used for weighting is
\begin{equation}
\Sigma_{C,i} \;=\; G_i\,\Sigma_{\mathrm{rel},i}\,G_i\transp .
\label{eq:supp_SigmaCi}
\end{equation}

\paragraph{Track information.} $\Sigma_{\mathrm{rel},i}=(H_i+\lambda I)^{-1}$ is obtained from the
query-seeded tracks of reference $i$ by Gauss--Newton accumulation of reprojection residuals,
$H_i=\sum_j J_j\transp W_j J_j$, where $J_j\in\Real^{2\times6}$ is the per-track pose Jacobian, and
$W_j=(c_j/\sigma_{\mathrm{pix}}^2)I_2$ weights track $j$ by its VGGT confidence $c_j$ over a robust
MAD residual scale $\sigma_{\mathrm{pix}}=\max(1.4826\,\mathrm{MAD}(\lVert r\rVert),\sigma_{\min})$.
For PSD safety the information eigenvalues are floored at $10^{-8}$ and the resulting
$\Sigma_{\mathrm{rel},i}$ eigenvalues capped at $10^{6}$ (the asymmetric clamp of
Sec.~\ref{sec:implementation_details}).

\paragraph{Common-mode scale variance.} The shared scale variance is estimated robustly from the
pairwise reference-center ratios, without SfM 3D map points:
\begin{equation}
s_{ab}=\frac{\lVert C_a^m-C_b^m\rVert}{\lVert C_a^v-C_b^v\rVert},\qquad
\mathrm{var}_s=\frac{\big(1.4826\,\mathrm{MAD}(s_{ab})\big)^2}{N_h},
\label{eq:supp_vars}
\end{equation}
for $N_h$ retained covariance-bearing hypotheses when at least two valid ratios exist; otherwise
$\mathrm{var}_s=0$.

\paragraph{Robust consensus and final score.} The map-frame center is fused by Student-$t$ IRLS over
the retained covariance-bearing hypotheses $\mathcal{U}_q$. Let $N_h=|\mathcal{U}_q|$,
$\mathrm{Info}_k=\Sigma_{C,k}^{-1}$, and $\omega_k=(\nu+3)/(\nu+m_k^2)$, where
$m_k^2=(\Ccons-C_{q,k})\transp\mathrm{Info}_k(\Ccons-C_{q,k})$ and $\nu=5$. This gives
$\Ccons=A^{-1}b$, with
$A=\sum_{k\in\mathcal{U}_q}\omega_k\mathrm{Info}_k$ and
$b=\sum_{k\in\mathcal{U}_q}\omega_k\mathrm{Info}_k C_{q,k}$, and the aggregate
inverse-information approximation $\Sigma_{\mathrm{agg}}=A^{-1}$. A reduced $\chi^2$ of the weighted
residuals,
$\chi^2_{\mathrm{red}}=\tfrac{1}{\,\mathrm{dof}\,}\sum_{k\in\mathcal{U}_q}\omega_k m_k^2$, with
$\mathrm{dof}=\max(3N_h-3,1)$, inflates this aggregate covariance approximation. The scale variance
is then added once as a common-mode aggregate term, yielding the scalar score
\begin{equation}
\sigcons \;=\; \sqrt{\;\lambda_{\max}\!\Big(\phi\,\Sigma_{\mathrm{agg}}
\;+\; \mathrm{var}_s\,\tfrac{\partial\Ccons}{\partial \sscale}\big(\tfrac{\partial\Ccons}{\partial \sscale}\big)\transp\Big)},
\qquad \phi=\max(1,\chi^2_{\mathrm{red}}).
\label{eq:supp_signew}
\end{equation}
This is an aggregate covariance approximation used for failure ranking, not a calibrated posterior,
probability, or confidence region.
The ablation comparator \sigdisp{} is the information-blind RMS dispersion
$\sigdisp=\sqrt{\tfrac{1}{N_h}\sum_{k\in\mathcal{U}_q}\lVert C_{q,k}-\Ccons\rVert^2}$. The
center-lift Jacobian is verified numerically by finite differences.

\paragraph{Calibration of the implied covariance.} \sigcons{} is used as a ranking score, not as a
calibrated covariance. \Cref{tab:cov_calib} quantifies this: standardizing the consensus-center error
by the predicted scale, $\lVert C_{\mathrm{gt}}-\Ccons\rVert/\sigcons$, has median $4.0$--$5.8$ (a
calibrated $3$-DoF Gaussian would give $\approx1.6$ using the largest-axis $\sigma$), and only
$35$--$40\%$ of true centers fall within the $3\sigcons$ radius (nominal $\approx97\%$). The
first-order, information-additive covariance is thus over-confident in absolute scale by
$\approx4$--$6\times$, as expected since the $N_h$ retained hypotheses share one VGGT pass, depth, and scale yet
are aggregated as if independent. This does not affect the selective use of \sigcons{}: as a monotone
failure-ranking signal it is effective (main-paper \MainCref{tab:rejecthead2head}; calibration-in-ranking
in \cref{sec:supp_calib}), which is why we present it as a ranking score rather than a posterior. A
rotation-reliability term and a properly correlated covariance are left to future work.

\begin{table}[t]\centering
\caption{Calibration of the consensus-center covariance: median standardized residual
$\lVert C_{\mathrm{gt}}-\Ccons\rVert/\sigcons$ and the fraction of true centers within $k\,\sigcons$
(eligible queries). The covariance is over-confident in absolute scale ($\sim4$--$6\times$),
consistent with its use as a ranking signal rather than a calibrated posterior.}
\label{tab:cov_calib}
\setlength{\tabcolsep}{6pt}
\begin{tabular}{lcccc}
\toprule
Dataset & median $\lVert e\rVert/\sigcons$ & $<\!1\sigcons$ & $<\!2\sigcons$ & $<\!3\sigcons$ \\
\midrule
7-Scenes  & 4.04 & 5.4\%  & 19.4\% & 35.2\% \\
Cambridge & 3.99 & 8.2\%  & 22.2\% & 37.1\% \\
NAVER-B2  & 5.78 & 27.0\% & 30.9\% & 39.7\% \\
NAVER-B1  & 5.30 & 23.3\% & 29.7\% & 36.3\% \\
\bottomrule
\end{tabular}
\end{table}

\section{Open-world selection without held-out normalization}
\label{sec:supp_openworld}

The main-paper \sigjoint{} uses leave-one-scene-out / cross-building held-out medians to align the
scales of \sigcons{} and \sigdisp{}. \Cref{tab:supp_openworld} shows this normalization is
optional: a fully held-out-free rank-based fusion
$\max(\mathrm{rank}(\sigcons),\mathrm{rank}(\sigdisp))$ matches the held-out joint within $0.5$\,pp on
7-Scenes/Cambridge, and on the low-texture NAVER buildings the single scale-free signal \sigcons{}
(AUROC is invariant to any monotone rescaling, hence needs no normalization at all) already exceeds
the held-out joint. These results suggest that a deployment with zero held-out data can use rank
fusion, or \sigcons{} alone in low-texture regimes, without relying on held-out scale normalization.

\begin{table}[t]\centering
\caption{Failure-detection AUROC under held-out-free selection. Rank fusion needs no normalization;
\sigcons{} alone is scale-free. Bold: best held-out-free entry per column.}
\label{tab:supp_openworld}
\setlength{\tabcolsep}{6pt}
\resizebox{\columnwidth}{!}{%
\begin{tabular}{lcccc}
\toprule
Selection score & 7-Scenes & Cambridge & NAVER-B2 & NAVER-B1 \\
\midrule
\sigjoint{} (held-out norm., main paper)      & 0.822 & 0.767 & 0.791 & 0.781 \\
\sigjoint{} (rank fusion, no held-out) & \textbf{0.818} & \textbf{0.768} & 0.748 & 0.766 \\
\sigcons{} alone (scale-free)                 & 0.778 & 0.764 & \textbf{0.806} & \textbf{0.813} \\
\bottomrule
\end{tabular}}
\end{table}

\section{Covariance-eligibility breakdown}
\label{sec:supp_elig}

A query is covariance-eligible when the covariance consensus produces finite reliability signals
from at least two covariance-bearing hypotheses, including inflated weak hypotheses when needed;
ineligible queries count as failures in full-set recall. \Cref{tab:supp_elig} records post-hoc
diagnostic annotations from the evaluation outputs. The listed same-space annotations describe many
failed cases, but they are not a separate eligibility gate in the covariance implementation.

\begin{table}[t]\centering
\caption{Covariance-ineligible queries and their dominant post-hoc diagnostic annotation.
``Same-space fail'' denotes a stored score-$0$ annotation, not an additional covariance gate.}
\label{tab:supp_elig}
\setlength{\tabcolsep}{6pt}
\begin{tabular}{lccl}
\toprule
Dataset & Eligible \% & Ineligible \% & Dominant cause \\
\midrule
7-Scenes   & 100.0 & 0.0  & --- \\
Cambridge  & 98.8  & 1.2  & same-space / align fail \\
NAVER-B2   & 47.3  & 52.7 & same-space fail (score $0$) \\
NAVER-B1   & 74.7  & 25.3 & same-space fail (score $0$) \\
\bottomrule
\end{tabular}
\end{table}

\section{Statistical significance and runtime}
\label{sec:supp_ci}

\Cref{tab:supp_ci} reports $95\%$ bootstrap confidence intervals ($2000$ query resamples) for the
held-out-free rank-fusion variant and strict-success rate; all AUROC intervals exclude the $0.5$
chance level. \Cref{tab:supp_runtime} gives the median per-stage runtime: the frozen VGGT forward
pass dominates, while reference selection and the consensus/covariance computation are negligible.

\begin{table}[t]\centering
\caption{Bootstrap $95\%$ CIs (2000 resamples over queries). Joint $=$ rank-fused \sigjoint{}.}
\label{tab:supp_ci}
\setlength{\tabcolsep}{5pt}
\begin{tabular}{lccc}
\toprule
Dataset & joint AUROC & \sigcons{} AUROC & strict success \\
\midrule
7-Scenes  & [0.811,\,0.825] & [0.769,\,0.786] & [0.759,\,0.772] \\
Cambridge & [0.744,\,0.791] & [0.742,\,0.787] & [0.294,\,0.335] \\
NAVER-B2  & [0.702,\,0.791] & [0.764,\,0.847] & [0.462,\,0.554] \\
NAVER-B1  & [0.744,\,0.786] & [0.794,\,0.832] & [0.521,\,0.564] \\
\bottomrule
\end{tabular}
\end{table}

\begin{table}[t]\centering
\caption{Median per-query runtime (ms) by stage. VGGT is the frozen forward pass. Per-stage values
are independent per-query medians and need not sum to the median total. ``refine'' times a
diagnostic PnP-refinement attempt whose output is never used for any reported pose; its cost is
included for completeness.}
\label{tab:supp_runtime}
\setlength{\tabcolsep}{5pt}
\begin{tabular}{lccccccc}
\toprule
Dataset & preproc. & retrieval & selection & VGGT & align & refine & total \\
\midrule
7-Scenes  & 23 & 33  & 17 & 1212 & 409 & 146 & 1859 \\
Cambridge & 38 & 86  & 14 & 1739 & 339 & 161 & 2428 \\
NAVER-B2  & 48 & 271 & 18 & 2074 & 862 & 0   & 3266 \\
NAVER-B1  & 43 & 665 & 10 & 2045 & 432 & 76  & 3429 \\
\bottomrule
\end{tabular}
\end{table}

\Cref{tab:significance} reports (i) the full-set selective AUROC, where covariance-ineligible queries
are ranked least-reliable and counted as failures (so the score reflects the whole query stream, not
just the eligible subset), and (ii) a paired bootstrap test of the joint-vs-\sigcons{} AUROC
difference on the eligible set ($2000$ resamples). The full-set convention places covariance-ineligible
queries last; the eligible fractions ($47.3\%$/$74.7\%$ on B2/B1) are the deployable coverage ceiling and are stated
alongside the eligible-set numbers in the main paper. The paired test shows the joint score's
advantage over \sigcons{} is significant only on 7-Scenes, not separable on Cambridge, and
significantly negative on both NAVER buildings, supporting \sigcons{} as the single-signal
default in low-texture regimes. ($\Delta$ uses the held-out-free rank fusion of
\cref{sec:supp_openworld}; the held-out joint of the main paper shows the same ordering.)

\begin{table}[t]\centering
\caption{Eligibility coverage, full-set selective AUROC (ineligible ranked least-reliable and counted
as failures), and paired bootstrap $95\%$ CI of the $\sigjoint-\sigcons$ AUROC difference on the
eligible set. ``*'' = CI excludes $0$ (significant); ``ns'' = not separable.}
\label{tab:significance}
\setlength{\tabcolsep}{5pt}
\begin{tabular}{lccc}
\toprule
Dataset & eligible \% & full-set AUROC ($\sigjoint$/$\sigcons$) & $\Delta(\sigjoint-\sigcons)$ 95\% CI \\
\midrule
7-Scenes  & 100.0 & 0.818 / 0.778 & $[+0.034,+0.047]$* \\
Cambridge & 98.8  & 0.772 / 0.768 & $[-0.012,+0.020]$ ns \\
NAVER-B2  & 47.3  & 0.923 / 0.941 & $[-0.087,-0.028]$* \\
NAVER-B1  & 74.7  & 0.865 / 0.893 & $[-0.062,-0.034]$* \\
\bottomrule
\end{tabular}
\end{table}

\Cref{tab:fullstream} isolates the failure-ranking value of \sigcons{} beyond the eligibility
filter. We compare against a gate-only ranker (rank purely by covariance eligibility, ineligible last)
on the full query stream, and against scalar dispersion \sigdisp{} under fixed eligibility. On the
full stream \sigcons{} exceeds the gate-only ranker on every dataset (paired bootstrap of the difference
is significant everywhere; e.g.\ NAVER-B1 $0.893$ vs.\ $0.713$), separating the primary gain from the
eligibility filter. Under fixed eligibility, \sigcons{} outperforms \sigdisp{} significantly on
7-Scenes/NAVER-B2/B1 (tied on Cambridge), confirming the covariance score adds discriminative value
over a simpler dispersion signal. Full-stream strict risk at $0.5$ coverage also drops well below the
gate-only ranker (e.g.\ NAVER-B1 $0.280$ vs.\ $0.460$; 7-Scenes $0.094$ vs.\ $0.234$).

\begin{table}[t]\centering
\caption{Failure-ranking value of \sigcons{} beyond the eligibility gate. Full-stream AUROC ranks all
queries (ineligible counted as failures, placed last); the gate-only ranker uses covariance
eligibility alone. Within-eligible columns hold eligibility fixed. $\Delta$ columns are paired
bootstrap $95\%$ CIs ($2000$ resamples); ``*'' excludes $0$, ``ns'' not separable.}
\label{tab:fullstream}
\setlength{\tabcolsep}{4pt}
\resizebox{\columnwidth}{!}{%
\begin{tabular}{lcccccc}
\toprule
 & \multicolumn{3}{c}{full-stream AUROC} & \multicolumn{1}{c}{within-elig.} & \multicolumn{2}{c}{$\Delta$ 95\% CI} \\
\cmidrule(lr){2-4}\cmidrule(lr){5-5}\cmidrule(lr){6-7}
Dataset & gate-only & $\sigcons$ & $\sigdisp$ & $\sigcons$/$\sigdisp$ & $\sigcons{-}$gate & $\sigcons{-}\sigdisp$ \\
\midrule
7-Scenes  & 0.500 & 0.778 & 0.762 & 0.778/0.762 & $[{+}0.27,{+}0.29]$* & $[{+}0.01,{+}0.03]$* \\
Cambridge & 0.509 & 0.768 & 0.745 & 0.764/0.741 & $[{+}0.24,{+}0.28]$* & $[{-}0.00,{+}0.05]$\,ns \\
NAVER-B2  & 0.847 & 0.941 & 0.883 & 0.806/0.618 & $[{+}0.08,{+}0.11]$* & $[{+}0.13,{+}0.24]$* \\
NAVER-B1  & 0.713 & 0.893 & 0.823 & 0.813/0.692 & $[{+}0.17,{+}0.19]$* & $[{+}0.10,{+}0.15]$* \\
\bottomrule
\end{tabular}}
\end{table}

\section{Empirical ranking behavior of \texorpdfstring{\sigcons}{sigma\_cons}}
\label{sec:supp_calib}

\sigcons{} is an approximate covariance-derived ranking signal, not a calibrated probability.
\Cref{tab:supp_calib} bins queries into \sigcons{} deciles (low to high). Failure rates generally
increase with \sigcons{}, with finite-sample nonmonotonicity on some splits, and Spearman
correlations remain positive ($+0.41$ to $+0.54$). The score is thus well-ordered for selective
ranking even though it is not a calibrated probability.

\begin{table}[t]\centering
\caption{Empirical strict-failure rate per \sigcons{} decile (low$\to$high) and Spearman correlation
with the failure indicator.}
\label{tab:supp_calib}
\setlength{\tabcolsep}{3.2pt}
\begin{tabular}{lccccccccccc}
\toprule
Dataset & D1 & D2 & D3 & D4 & D5 & D6 & D7 & D8 & D9 & D10 & Spearman \\
\midrule
7-Scenes  & .03 & .09 & .09 & .11 & .15 & .19 & .25 & .31 & .47 & .65 & $+.41$ \\
Cambridge & .30 & .49 & .54 & .58 & .64 & .74 & .85 & .86 & .93 & .94 & $+.43$ \\
NAVER-B2  & .18 & .25 & .30 & .16 & .28 & .47 & .81 & .65 & .84 & 1.0 & $+.53$ \\
NAVER-B1  & .17 & .17 & .22 & .24 & .30 & .39 & .56 & .66 & .86 & .99 & $+.54$ \\
\bottomrule
\end{tabular}
\end{table}

\section{Naive consensus-inlier-count baseline}
\label{sec:supp_inlier}

A natural ground-truth-free confidence is the size of the robust consensus (the analog of PnP/RANSAC
inlier count). \Cref{tab:supp_inlier} shows this is near-random for pose-failure ranking (AUROC
$0.50$--$0.56$): with $K{=}8$ hypotheses the inlier count takes only values in $\{5,6,7,8\}$ and
rarely discriminates. This helps explain why the propagated covariance score \sigcons{}
($0.76$--$0.81$) provides a stronger failure-ranking signal.

\begin{table}[t]\centering
\caption{Consensus-inlier-count as a failure-detection signal (scored in its more informative
orientation), versus \sigcons{}.}
\label{tab:supp_inlier}
\setlength{\tabcolsep}{6pt}
\begin{tabular}{lcccc}
\toprule
Signal & 7-Scenes & Cambridge & NAVER-B2 & NAVER-B1 \\
\midrule
consensus-inlier count & 0.501 & 0.507 & 0.503 & 0.564 \\
\sigcons{} (ours)      & \textbf{0.778} & \textbf{0.764} & \textbf{0.806} & \textbf{0.813} \\
\bottomrule
\end{tabular}
\end{table}

\section{Per-scene 12-Scenes results}
\label{sec:supp_12scenes}

\Cref{tab:supp_12scenes} expands the 12-Scenes block of the main paper to all twelve scenes
(median translation in cm / strict success at $5\,\text{cm}/5^{\circ}$). \method{} achieves higher
strict success than the comparable SfM-point-map-free Reloc3r-512 baseline on every scene and lower
median translation on most scenes; the structure-based hloc upper bound is shown for context.

\begin{table}[t]\centering
\caption{Per-scene 12-Scenes~\cite{twelvescenes2016} accuracy (median $t$ cm / strict@$5\,\text{cm}/5^{\circ}$).
hloc is structure-based (context, not like-for-like); \method{} and Reloc3r-512 use no precomputed
SfM 3D point map in this comparison.}
\label{tab:supp_12scenes}
\setlength{\tabcolsep}{5pt}
\begin{tabular}{lccc}
\toprule
Scene & \method{} & hloc & Reloc3r-512 \\
\midrule
apt1\_kitchen      & 1.34 / 0.982 & 0.66 / 1.000 & 3.25 / 0.560 \\
apt1\_living       & 1.34 / 1.000 & 0.92 / 1.000 & 1.94 / 0.787 \\
apt2\_bed          & 2.06 / 0.939 & 0.94 / 0.991 & 12.28 / 0.237 \\
apt2\_kitchen      & 1.29 / 0.979 & 0.76 / 1.000 & 2.34 / 0.708 \\
apt2\_living       & 2.08 / 0.791 & 1.13 / 0.994 & 7.47 / 0.406 \\
apt2\_luke         & 1.57 / 0.920 & 1.36 / 0.972 & 3.71 / 0.649 \\
office1\_gates362  & 1.69 / 0.985 & 1.24 / 1.000 & 1.64 / 0.975 \\
office1\_gates381  & 0.62 / 0.968 & 0.67 / 0.975 & 1.46 / 0.948 \\
office1\_lounge    & 2.32 / 0.886 & 1.62 / 0.971 & 3.32 / 0.681 \\
office1\_manolis   & 1.25 / 0.983 & 0.80 / 1.000 & 2.37 / 0.807 \\
office2\_5a        & 1.39 / 0.940 & 1.32 / 0.912 & 2.53 / 0.730 \\
office2\_5b        & 1.40 / 0.980 & 1.01 / 1.000 & 3.22 / 0.704 \\
\midrule
\textbf{scene mean} & 1.53 / 0.946 & 1.04 / 0.985 & 3.79 / 0.683 \\
\textbf{pooled ($5641$ q)} & 1.33 / 0.957 & 0.95 / 0.985 & 2.32 / 0.750 \\
\bottomrule
\end{tabular}
\end{table}

\clearpage
\bibliographystyle{splncs04}
\bibliography{references}

\begin{thebibliography}{10}
\providecommand{\url}[1]{\texttt{#1}}
\providecommand{\urlprefix}{URL }
\providecommand{\doi}[1]{https://doi.org/#1}

\bibitem{angelopoulos2023gentle}
Angelopoulos, A.N., Bates, S.: Conformal prediction: A gentle introduction.
  Foundations and Trends in Machine Learning  \textbf{16}(4),  494--591 (03
  2023). \doi{10.1561/2200000101}, \url{https://doi.org/10.1561/2200000101}

\bibitem{learnthentest2025}
Angelopoulos, A.N., Bates, S., Cand{\`e}s, E.J., Jordan, M.I., Lei, L.: Learn
  then test: Calibrating predictive algorithms to achieve risk control. The
  Annals of Applied Statistics  \textbf{19}(2),  1641--1662 (2025)

\bibitem{netvlad2016}
Arandjelovic, R., Gronat, P., Torii, A., Pajdla, T., Sivic, J.: Netvlad: Cnn
  architecture for weakly supervised place recognition. In: 2016 IEEE
  Conference on Computer Vision and Pattern Recognition (CVPR). pp. 5297--5307
  (2016). \doi{10.1109/CVPR.2016.572}

\bibitem{mickey2024}
Barroso-Laguna, A., Munukutla, S., Prisacariu, V.A., Brachmann, E.: Matching 2d
  images in 3d: Metric relative pose from metric correspondences. In: 2024
  IEEE/CVF Conference on Computer Vision and Pattern Recognition (CVPR). pp.
  4852--4863 (2024). \doi{10.1109/CVPR52733.2024.00464}

\bibitem{megaloc2025}
Berton, G., Masone, C.: Megaloc: One retrieval to place them all. In: 2025
  IEEE/CVF Conference on Computer Vision and Pattern Recognition Workshops
  (CVPRW). pp. 2852--2858 (2025). \doi{10.1109/CVPRW67362.2025.00269}

\bibitem{ace2023}
Brachmann, E., Cavallari, T., Prisacariu, V.A.: Accelerated coordinate
  encoding: Learning to relocalize in minutes using rgb and poses. In: 2023
  IEEE/CVF Conference on Computer Vision and Pattern Recognition (CVPR). pp.
  5044--5053 (2023). \doi{10.1109/CVPR52729.2023.00488}

\bibitem{dsac2017}
Brachmann, E., Krull, A., Nowozin, S., Shotton, J., Michel, F., Gumhold, S.,
  Rother, C.: Dsac — differentiable ransac for camera localization. In: 2017
  IEEE Conference on Computer Vision and Pattern Recognition (CVPR). pp.
  2492--2500 (2017). \doi{10.1109/CVPR.2017.267}

\bibitem{geomfactors2018}
Brahmbhatt, S., Gu, J., Kim, K., Hays, J., Kautz, J.: Geometry-aware learning
  of maps for camera localization. In: 2018 IEEE/CVF Conference on Computer
  Vision and Pattern Recognition. pp. 2616--2625 (2018).
  \doi{10.1109/CVPR.2018.00277}

\bibitem{marepo2024}
Chen, S., Cavallari, T., Prisacariu, V.A., Brachmann, E.: Map-relative pose
  regression for visual re-localization. In: 2024 IEEE/CVF Conference on
  Computer Vision and Pattern Recognition (CVPR). pp. 20665--20674 (2024).
  \doi{10.1109/CVPR52733.2024.01953}

\bibitem{superpoint2018}
DeTone, D., Malisiewicz, T., Rabinovich, A.: Superpoint: Self-supervised
  interest point detection and description. In: 2018 IEEE/CVF Conference on
  Computer Vision and Pattern Recognition Workshops (CVPRW). pp. 337--33712
  (2018). \doi{10.1109/CVPRW.2018.00060}

\bibitem{reloc3r2025}
Dong, S., Wang, S., Liu, S., Cai, L., Fan, Q., Kannala, J., Yang, Y.: Reloc3r:
  Large-scale training of relative camera pose regression for generalizable,
  fast, and accurate visual localization. In: 2025 IEEE/CVF Conference on
  Computer Vision and Pattern Recognition (CVPR). pp. 16739--16752 (2025).
  \doi{10.1109/CVPR52734.2025.01560}

\bibitem{elyaniv2010}
El-Yaniv, R., Wiener, Y.: On the foundations of noise-free selective
  classification. Journal of Machine Learning Research  \textbf{11}(53),
  1605--1641 (2010), \url{http://jmlr.org/papers/v11/el-yaniv10a.html}

\bibitem{ransac1981}
Fischler, M.A., Bolles, R.C.: Random sample consensus: a paradigm for model
  fitting with applications to image analysis and automated cartography.
  Commun. ACM  \textbf{24}(6),  381–395 (Jun 1981).
  \doi{10.1145/358669.358692}, \url{https://doi.org/10.1145/358669.358692}

\bibitem{govindu2001}
Govindu, V.: Combining two-view constraints for motion estimation. In:
  Proceedings of the 2001 IEEE Computer Society Conference on Computer Vision
  and Pattern Recognition. CVPR 2001. vol.~2, pp. II--II (2001).
  \doi{10.1109/CVPR.2001.990963}

\bibitem{rotavg2013}
Hartley, R., Trumpf, J., Dai, Y., Li, H.: Rotation averaging. International
  journal of computer vision  \textbf{103}(3),  267--305 (2013)

\bibitem{bayesianposenet2016}
Kendall, A., Cipolla, R.: Modelling uncertainty in deep learning for camera
  relocalization. In: 2016 IEEE International Conference on Robotics and
  Automation (ICRA). pp. 4762--4769 (2016). \doi{10.1109/ICRA.2016.7487679}

\bibitem{posenet2015}
Kendall, A., Grimes, M., Cipolla, R.: Posenet: A convolutional network for
  real-time 6-dof camera relocalization. In: 2015 IEEE International Conference
  on Computer Vision (ICCV). pp. 2938--2946 (2015). \doi{10.1109/ICCV.2015.336}

\bibitem{naver2021}
Lee, D., Ryu, S., Yeon, S., Lee, Y., Kim, D., Han, C., Cabon, Y., Weinzaepfel,
  P., Guerin, N., Csurka, G., Humenberger, M.: { Large-scale Localization
  Datasets in Crowded Indoor Spaces }. In: 2021 IEEE/CVF Conference on Computer
  Vision and Pattern Recognition (CVPR). pp. 3226--3235. IEEE Computer Society,
  Los Alamitos, CA, USA (Jun 2021). \doi{10.1109/CVPR46437.2021.00324}

\bibitem{mast3r2024}
Leroy, V., Cabon, Y., Revaud, J.: Grounding image matching in 3d with mast3r.
  In: European conference on computer vision. pp. 71--91. Springer (2024)

\bibitem{lightglue2023}
Lindenberger, P., Sarlin, P.E., Pollefeys, M.: Lightglue: Local feature
  matching at light speed. In: 2023 IEEE/CVF International Conference on
  Computer Vision (ICCV). pp. 17581--17592 (2023).
  \doi{10.1109/ICCV51070.2023.01616}

\bibitem{robotcar2017}
Maddern, W., Pascoe, G., Linegar, C., Newman, P.: 1 year, 1000 km: The oxford
  robotcar dataset. The International Journal of Robotics Research
  \textbf{36}(1),  3--15 (2017)

\bibitem{dinov2_2024}
Oquab, M., Darcet, T., Moutakanni, T., Vo, H.V., Szafraniec, M., Khalidov, V.,
  Fernandez, P., HAZIZA, D., Massa, F., El-Nouby, A., Assran, M., Ballas, N.,
  Galuba, W., Howes, R., Huang, P.Y., Li, S.W., Misra, I., Rabbat, M., Sharma,
  V., Synnaeve, G., Xu, H., Jegou, H., Mairal, J., Labatut, P., Joulin, A.,
  Bojanowski, P.: {DINO}v2: Learning robust visual features without
  supervision. Transactions on Machine Learning Research  (2024),
  \url{https://openreview.net/forum?id=a68SUt6zFt}, featured Certification

\bibitem{hloc2019}
Sarlin, P.E., Cadena, C., Siegwart, R., Dymczyk, M.: From coarse to fine:
  Robust hierarchical localization at large scale. In: 2019 IEEE/CVF Conference
  on Computer Vision and Pattern Recognition (CVPR). pp. 12708--12717 (2019).
  \doi{10.1109/CVPR.2019.01300}

\bibitem{superglue2020}
Sarlin, P.E., DeTone, D., Malisiewicz, T., Rabinovich, A.: Superglue: Learning
  feature matching with graph neural networks. In: 2020 IEEE/CVF Conference on
  Computer Vision and Pattern Recognition (CVPR). pp. 4937--4946 (2020).
  \doi{10.1109/CVPR42600.2020.00499}

\bibitem{pixloc2021}
Sarlin, P.E., Unagar, A., Larsson, M., Germain, H., Toft, C., Larsson, V.,
  Pollefeys, M., Lepetit, V., Hammarstrand, L., Kahl, F., Sattler, T.: Back to
  the feature: Learning robust camera localization from pixels to pose. In:
  2021 IEEE/CVF Conference on Computer Vision and Pattern Recognition (CVPR).
  pp. 3246--3256 (2021). \doi{10.1109/CVPR46437.2021.00326}

\bibitem{activesearch2016}
Sattler, T., Leibe, B., Kobbelt, L.: Efficient \& effective prioritized
  matching for large-scale image-based localization. IEEE Transactions on
  Pattern Analysis and Machine Intelligence  \textbf{39}(9),  1744--1756
  (2017). \doi{10.1109/TPAMI.2016.2611662}

\bibitem{aachen2018}
Sattler, T., Maddern, W., Toft, C., Torii, A., Hammarstrand, L., Stenborg, E.,
  Safari, D., Okutomi, M., Pollefeys, M., Sivic, J., Kahl, F., Pajdla, T.:
  Benchmarking 6dof outdoor visual localization in changing conditions. In:
  2018 IEEE/CVF Conference on Computer Vision and Pattern Recognition. pp.
  8601--8610 (2018). \doi{10.1109/CVPR.2018.00897}

\bibitem{colmap2016}
Schönberger, J.L., Frahm, J.M.: Structure-from-motion revisited. In: 2016 IEEE
  Conference on Computer Vision and Pattern Recognition (CVPR). pp. 4104--4113
  (2016). \doi{10.1109/CVPR.2016.445}

\bibitem{sevenscenes2013}
Shotton, J., Glocker, B., Zach, C., Izadi, S., Criminisi, A., Fitzgibbon, A.:
  Scene coordinate regression forests for camera relocalization in rgb-d
  images. In: 2013 IEEE Conference on Computer Vision and Pattern Recognition.
  pp. 2930--2937 (2013). \doi{10.1109/CVPR.2013.377}

\bibitem{inloc2018}
Taira, H., Okutomi, M., Sattler, T., Cimpoi, M., Pollefeys, M., Sivic, J.,
  Pajdla, T., Torii, A.: Inloc: Indoor visual localization with dense matching
  and view synthesis. IEEE Transactions on Pattern Analysis and Machine
  Intelligence  \textbf{43}(4),  1293--1307 (2021).
  \doi{10.1109/TPAMI.2019.2952114}

\bibitem{umeyama1991}
Umeyama, S.: Least-squares estimation of transformation parameters between two
  point patterns. IEEE Transactions on Pattern Analysis and Machine
  Intelligence  \textbf{13}(4),  376--380 (1991). \doi{10.1109/34.88573}

\bibitem{twelvescenes2016}
Valentin, J., Dai, A., Niessner, M., Kohli, P., Torr, P., Izadi, S., Keskin,
  C.: Learning to navigate the energy landscape. In: 2016 Fourth International
  Conference on 3D Vision (3DV). pp. 323--332 (2016). \doi{10.1109/3DV.2016.41}

\bibitem{glace2024}
Wang, F., Jiang, X., Galliani, S., Vogel, C., Pollefeys, M.: { GLACE: Global
  Local Accelerated Coordinate Encoding }. In: 2024 IEEE/CVF Conference on
  Computer Vision and Pattern Recognition (CVPR). pp. 5819--5828. IEEE Computer
  Society, Los Alamitos, CA, USA (Jun 2024).
  \doi{10.1109/CVPR52733.2024.02037},
  \url{https://doi.ieeecomputersociety.org/10.1109/CVPR52733.2024.02037}

\bibitem{vggt2025}
Wang, J., Chen, M., Karaev, N., Vedaldi, A., Rupprecht, C., Novotny, D.: Vggt:
  Visual geometry grounded transformer. In: 2025 IEEE/CVF Conference on
  Computer Vision and Pattern Recognition (CVPR). pp. 5294--5306 (2025).
  \doi{10.1109/CVPR52734.2025.00499}

\bibitem{dust3r2024}
Wang, S., Leroy, V., Cabon, Y., Chidlovskii, B., Revaud, J.: Dust3r: Geometric
  3d vision made easy. In: 2024 IEEE/CVF Conference on Computer Vision and
  Pattern Recognition (CVPR). pp. 20697--20709 (2024).
  \doi{10.1109/CVPR52733.2024.01956}

\end{thebibliography}

\end{document}